\documentclass{article}

\usepackage{microtype}
\usepackage{graphicx}
\usepackage{subcaption}
\usepackage{booktabs} 

\usepackage{hyperref}
\usepackage{multirow}
\usepackage{array}
\usepackage{adjustbox}

\usepackage[preprint]{icml2026}

\usepackage{amsmath}
\usepackage{amssymb}
\usepackage{mathtools}
\usepackage{amsthm}

\usepackage[capitalize,noabbrev]{cleveref}

\usepackage{comment}
\usepackage{dirtytalk}
\usepackage{multirow}
\usepackage{array}
\usepackage{amsfonts}
\usepackage{amsmath}
\usepackage{arydshln}
\usepackage{fancyvrb}
\usepackage{fvextra}
\usepackage{threeparttablex}
\usepackage{makecell}   
\usepackage{caption}
\newcommand{\pb}[1]{\vspace{0.75ex}\noindent{\bf \em #1}\hspace*{.3em}}

\theoremstyle{plain}
\newtheorem{theorem}{Theorem}[section]
\newtheorem{proposition}[theorem]{Proposition}

\theoremstyle{definition}

\theoremstyle{remark}

\usepackage{algorithm}
\usepackage{algorithmic}
\usepackage{amsmath} 
\usepackage{amssymb}
\usepackage{booktabs}
\usepackage{multirow}
\usepackage{xcolor}
\usepackage{svg}
\usepackage[textsize=tiny]{todonotes}
\usepackage[table]{xcolor}

\icmltitlerunning{Geometric Neural Operators via Lie Group-Constrained Latent Dynamics}

\begin{document}

\twocolumn[
  \icmltitle{Geometric Neural Operators via Lie Group-Constrained Latent Dynamics}



  \icmlsetsymbol{equal}{*}

  \begin{icmlauthorlist}
    \icmlauthor{Jiaquan Zhang}{1}
    \icmlauthor{Fachrina Dewi Puspitasari}{2}
    \icmlauthor{Songbo Zhang}{1}
    \icmlauthor{Yibei Liu}{1}
    \icmlauthor{Kuien Liu}{3}
    \icmlauthor{Caiyan Qin}{4}
    \icmlauthor{Fan Mo}{5}
    \icmlauthor{Peng Wang}{2}
    \icmlauthor{Yang Yang}{2}
    \icmlauthor{Chaoning Zhang}{2}
  \end{icmlauthorlist}

  \icmlaffiliation{1}{School of Information and Software Engineering, University of Electronic Science and Technology of China, Chengdu, China}
  \icmlaffiliation{2}{Computer Science and Engineering, University of Electronic Science and Technology of China, Chengdu, China}
  \icmlaffiliation{3}{Institute of Software Chinese Academy of Sciences, Beijing, China}
  \icmlaffiliation{4}{School of Robotics and Advanced Manufacture, Harbin Institute of Technology, Shenzhen, China}
  \icmlaffiliation{5}{Department of Computer Science, University of Oxford, Oxford, United Kingdom}

  \icmlcorrespondingauthor{Chaoning Zhang}{chaoningzhang@uestc.edu.cn}


  \vskip 0.3in
]



\printAffiliationsAndNotice{}  

\begin{abstract}
Neural operators offer an effective framework for learning solutions of partial differential equations for many physical systems in a resolution-invariant and data-driven manner.
Existing neural operators, however, often suffer from instability in multi-layer iteration and long-horizon rollout, which stems from the unconstrained Euclidean latent space updates that violate the geometric and conservation laws.
To address this challenge, we propose to constrain manifolds with low-rank Lie algebra parameterization that performs group action updates on the latent representation.
Our method, termed Manifold Constraining based on Lie group (MCL), acts as an efficient \emph{plug-and-play} module that enforces geometric inductive bias to existing neural operators.
Extensive experiments on various partial differential equations, such as 1-D Burgers and 2-D Navier-Stokes, over a wide range of parameters and steps demonstrate that our method effectively lowers the relative prediction error by 30-50\% at the cost of 2.26\% of parameter increase.
The results show that our approach provides a scalable solution for improving long-term prediction fidelity by addressing the principled geometric constraints absent in the neural operator updates.

\end{abstract}

\section{Introduction}
Neural operators provide an attractive data-driven solution for partial differential equations (PDEs) and complex physical system modeling through direct function mapping~\cite{azizzadenesheli2024neural, cuomo2022scientific, li2024physics}.
Unlike conventional numerical method that rely on explicit discretization and manual operator construction, neural operators model global input-output dependencies in a continuous function space.
This mechanism enables generalization under various parameters, initial conditions, and function dimensions~\cite{zhao2025diffeomorphism}.

Fourier Neural Operator~\cite{li2020fourier} is a seminal work in neural operator modeling that introduces frequency domain processing of PDEs to capture global features, thus enabling generalization effectively.
Since then, several other works have made significant progress in improving the performance of neural operator modeling through various extension, such as hierarchical~\cite{rahman2022u} and graph~\cite{li2020geometry} representations.
However, despite their excellent performance, these models still struggle to maintain the prediction stability over multi-layer iteration and long-term prediction~\cite{michalowska2024neural}.
The rollout nature of the prediction naturally propagates the instabilities arise in the early layers and steps.
This leads to numerical divergence, which, in a physical system, a slight distortion may deem the prediction inaccurate.
This phenomenon is particularly evident in nonlinear systems.

\begin{figure}
    \centering
    \includegraphics[width=\linewidth]{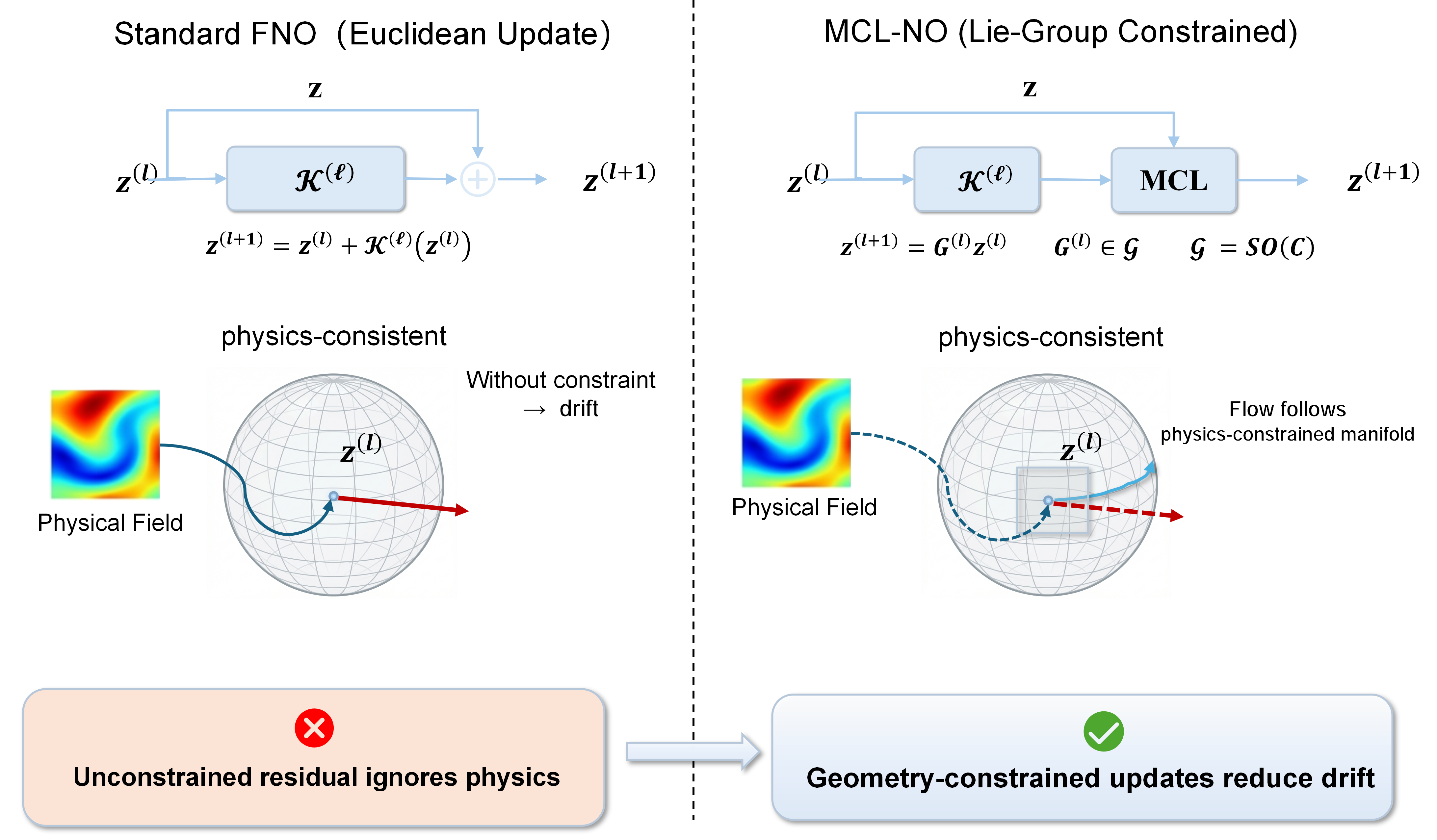}
    \caption{Conceptual comparison between common residual updates in existing neural operators vs. our geometric constrained updates. Our method, termed MCL, addresses the challenge of physical law violation in neural operator update by enforcing geometric inductive bias through Lie algebra parameterization.}
    \label{fig:teaser}
\end{figure}

The challenge of prediction instability in neural operators potentially stems from the absence of geometric constraints in the latent representation updates.
Geometric constraint refers to constraining model prediction on the natural characteristics of the PDE function, which is a substantial element in physical system modeling~\cite{viola2025pde}.
Nevertheless, conventional neural operators typically apply a residual connection commonly used in general neural networks to update the latent representation in Euclidean space.
The representation updates by this residual connection are mathematically similar to performing explicit unconstrained Euler integration on implicit dynamic systems~\cite{gracyk2025ricciflowregularizationlatent}.
These characteristics lack geometric constraining and make the latent representation updates prone to latent drift, a phenomenon that we observe as error propagation during inter-layer iteration, which leads to feature norm expansion and violation of physical consistency in existing neural operators.

Prior research has attempted to address this instability issue from three directions.
First, enhancing the frequency domain modeling with a multi-scale representation to improve model expressiveness on complex patterns~\cite{rahman2022u, liu2022ht}.
Second, improving network structure or introducing additional regularization loss to alleviate instability during training~\cite{li2024physics, runkel2024operator}.
Third, penalizing model prediction results with physics constraints or conservation terms~\cite{bulte2024probabilistic}.
These improvements, however, address the issue from the representation and optimization levels, instead of fixing the geometric structure modeling in the latent space.

Given that real physical system evolution naturally follows a well-defined geometric structure, we propose to introduce geometric inductive bias into the neural operators' latent representations' update mechanism.
Specifically, we propose a method termed MCL to address the latent drift issue in the neural operator due to a lack of geometric constraints during updates.
A Lie group~\cite{knapp1996lie} refers to a group of continuous algebraic transformations that forms a smooth manifold to capture its underlying geometric structure via differentiation.
As illustrated in Figure~\ref{fig:teaser}, our work implements the inductive bias through Lie group-structured hidden representations in the inter-layer update of neural operators.
Specifically, we replace the conventional Euclidean residual update with a group action update parameterized by Lie algebra to enforce hidden state evolution that obeys isometry and norm conservation.
Our proposed MCL is a \textit{plug-and-play} module that can be integrated into existing neural operators to improve their long-term stability and physical consistency.
To circumvent the computational bottleneck from the integration of MCL to the neural operator, we utilize low-rank approximations of the Lie Algebra and linearize the step sizes.
We summarize our contribution as follows.

\begin{itemize}
    \item We address the issue of latent drift in physical system evolution in the neural operator by enforcing geometric constraints to naturally follow real physical systems' behavior.
    \item We propose geometric inductive bias based on low-rank Lie-group structured representations that serves as an efficient plug-and-play module for neural operators.
    \item Through extensive experiments on various PDE equations, we demonstrate that our method significantly reduces rollout relative prediction errors (on average, 50\% reduction compared to baseline) at the cost of only 2.26\% parameter increases.
\end{itemize}

\section{Related Work}
\pb{Neural Operator.}
Neural operators aim to learn mappings between infinite-dimensional function spaces.
The models have become a prominent framework for solving PDEs in a resolution-invariant manner. 
The seminal Fourier Neural Operator (FNO)~\cite{li2020fourier} introduced spectral convolution to efficiently model global interactions in PDE solutions and demonstrated strong generalization across discretizations and parameter regimes.
Building on this idea, several variants have been proposed to improve expressiveness and geometric flexibility, including Graph Kernel Network (GKN)~\cite{li2020neural} for irregular meshes, Geometry-Informed Neural Operators (GINO)~\cite{li2020geometry} for complex domains, and U-shaped Neural Operators (U-NO)~\cite{rahman2022u} that incorporate multiscale encoder-decoder structures.
Other approaches, such as Deep Operator Networks (DeepONets)~\cite{lu2021learning}, learn operators via branch-trunk architectures and provide theoretical guarantees on universal approximation. 
Neural operators have also been extended to spatiotemporal settings for learning evolution operators in dynamic systems~\cite{hu2025deepomamba}. Recent works have also explored operator learning under a limited data regime~\cite{chen2024data}.

\pb{Geometric Constraints in Neural Operators.}
Latent drift is one of the core challenges in neural operators for physical systems, where learned dynamics can diverge from true PDE solutions.
A growing body of work seeks to mitigate this issue by embedding structural or temporal modeling principles.
Nonlocal Kernel Network (NKN)~\cite{you2022nonlocal} frames the operator update as a parabolic nonlocal equation to provide theoretical stability in the deep-layer limit.  
State-space neural operators explicitly model long-range dependencies using state-space formulations~\cite{hu2025state}. 
Recurrent Neural Operators (RNOs)~\cite{ye2025recurrent} directly integrates temporal recurrence into the operator to reduce cumulative error. 
GeoMaNO~\cite{han2025geomano} explores geometric and multiscale enhancements to neural operators to capture spatial structure during evolution. 
Additionally, aligning learned operators with long-term statistical properties of the underlying PDEs can help suppress divergence in chaotic regimes~\cite{jiang2023training}. 
Other works include physics-informed operator learning \cite{wang2021learning} and operator learning with conservation laws \cite{brandstetter2022message}.
Additionally, transformer-based operator models have been explored to better capture nonlocal dependencies \cite{cao2021choose}.
Despite these advances in structural, temporal, and invariant-preserving designs, the latent drift phenomenon that causes unstable long-term predictions remains the core bottleneck in neural operators.

\section{Method}
\subsection{Problem Statement}
A neural operator learns to map one function space $\mathcal{X}$ to another function space $\mathcal{Y}$.
Mathematically, it is defined as $\mathcal{G}: \mathcal{X} \rightarrow \mathcal{Y}$, where $u \in \mathcal{X}$ and $v \in \mathcal{Y}$ are continuous functions.
Existing neural operators typically learns function transformation through an iterative update mechanism in a finite-dimensional vector space by introducing latent space feature representations on a discrete grid.
Specifically, the neural operator takes as input hidden state $z^{(0)} \in \mathbb{R}^{C \times N}$, where $C$ and $N$ denote the channel dimension and number of discrete sampling points.
It then adopts the following inter-layer update form throughout $L$ layers of the neural network:
\begin{equation}\label{eq:update}
    z^{(l+1)} = z^{(l)} + \mathcal{K}^{(l)}\left(z^{(l)}\right), \quad l=0,\dots,L-1,
\end{equation}
where $\mathcal{K}^{(l)}$ represents the operator mapping of the $l$ layer. 
This update function is equivalent to performing explicit Euler integration on an implicit dynamic system in the Euclidean space.

\subsection{Motivation}
\begin{figure*}
    \centering
    \includegraphics[width=\linewidth]{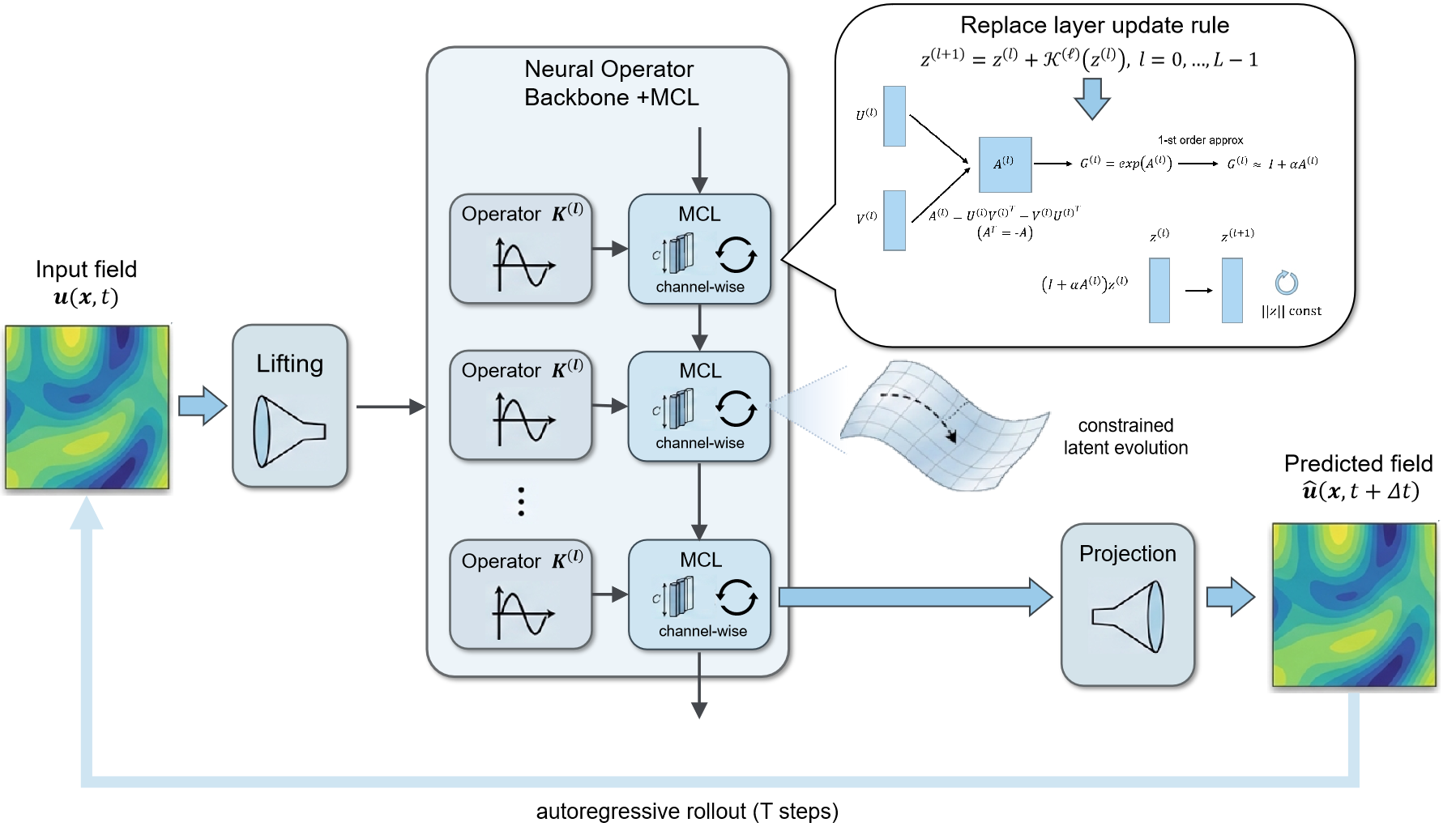}
    \caption{Framework of our method, termed MCL, when integrated in a neural operator. MCL replaces the standard unconstrained Euclidean space-based update with geometric-constrained updates based on Lie algebra group action. MCL serves as an efficient \emph{plug-and-play} module for existing neural operators.}
    \label{fig:framework}
\end{figure*}

The above update form, while being properly expressive, fails to account for the structural evolution of latent space features.
A real physical system is generally confined to a low-dimensional manifold or follows a specific geometric structure.
Updating latent representations solely with Equation~\ref{eq:update} makes them inherently prone to inter-layer deviation from the physically feasible subspace.
This phenomenon is known as latent drift, which accumulates during inter-layer iteration, leading to feature norms expansion and violation of physical consistency. 
To suppress this phenomenon, as illustrated in Figure~\ref{fig:framework}, we introduce a latent space update mechanism with geometric constraints that restrict the inter-layer feature evolution to a manifold with a well-defined geometric structure.

\subsection{Latent Space Update with Lie Group Constraints}
\pb{Lie Group Update.}
We assume that the inter-layer evolution of latent space features is generated by Lie group~\cite{knapp1996lie} actions, rather than vector addition in Euclidean space. 
Specifically, the update from layer $l$ to layer $l+1$ is defined as follows.
\begin{equation}
    z^{(l+1)} = G^{(l)} z^{(l)}, \quad G^{(l)} \in \mathcal{G},
\end{equation}
where $\mathcal{G}$ represents the selected Lie group. 
To ensure the isometry and numerical stability of the latent space features, this paper selects the orthogonal group $\mathcal{G} = \mathfrak{so}(C)$, consisting of all orthogonal transformations $G$ that satisfy $G^\top G = I$.

\pb{Lie Algebra Parameterization.} 
Since directly optimizing group elements $G^{(l)}$ is challenging, we instead parameterize the corresponding Lie algebra $\mathfrak{so}(C)$.
Lie algebra consists of all skew-symmetric matrices as follows.
\begin{equation}
    \mathfrak{so}(C) = \left\{ A \in \mathbb{R}^{C \times C} \mid A^\top = -A \right\}
\end{equation}
Given the generators $A^{(l)} \in \mathfrak{so}(C)$, group elements can be obtained through the exponential mapping.
\begin{equation}
    G^{(l)} = \exp\left(A^{(l)}\right)
\end{equation}
Since $\exp(A)$ always generates an orthogonal matrix for anti-symmetric matrices, the latent space update naturally satisfies norm conservation. 
\begin{equation}
    \|z^{(l+1)}\|_2 = \|z^{(l)}\|_2
\end{equation}

\subsection{Efficient Implementation with Low-rank Structure}
\pb{Low-rank Lie Algebra Approximation.}
Directly constructing a complete $C \times C$ skew-symmetric matrix is computationally and storage-wise expensive. 
To address this, we introduce a low-rank Lie algebra approximation to construct generators via two low-rank basis matrices $U^{(l)}, V^{(l)} \in \mathbb{R}^{C \times r}$ as follows.
\begin{equation}\label{eq:low-rank}
    A^{(l)} = U^{(l)} V^{(l)\top} - V^{(l)} U^{(l)\top},
\end{equation}
where $r \ll C$. 
This construction naturally satisfies anti-symmetry $A^{(l)\top} = -A^{(l)}.$
The low-rank form indicates that the latent space update only rotates within a low-dimensional subspace.
This is consistent with the assumption that the solution space of a physical system typically has a low-dimensional structure. 

\pb{Linearized Manifold Stepping.}
To further reduce computational complexity and simplify backpropagation, we adopt the first-order approximation form updated by Lie group. 
For small step sizes $\alpha > 0$, we obtain the following.
\begin{equation}
    \exp(\alpha A^{(l)}) \approx I + \alpha A^{(l)}
\end{equation}
Therefore, the update form can be written as follows.
\begin{equation}\label{eq:update}
    z^{(l+1)} = z^{(l)} + \alpha A^{(l)} z^{(l)}
\end{equation}
Although this form no longer strictly maintains orthogonality due to the anti-symmetric structure $A^{(l)}$, it still maintains a higher-order small quantity of norm change under first-order approximation.
This effectively suppresses latent space drift in neural operator updates. 
Implementing the low-rank approximation from Eq.~\ref{eq:low-rank}, the $A^{(l)} z^{(l)}$ part in Eq.~\ref{eq:update} can be expanded as follows.
\begin{equation}
A^{(l)} z = U^{(l)}\left(V^{(l)\top} z\right) - V^{(l)}\left(U^{(l)\top} z\right)
\end{equation}
This form reduces the computational complexity from $O(C^2)$ to $O(Cr)$ via batch matrix multiplication or einsum.

\pb{Integration with Neural Operators.}
The proposed Manifold-Constrained Layer (MCL) can be connected in series with any neural operator layer $\mathcal{K}$ as follows. 
\begin{equation}
\tilde z^{(l)} = \mathcal{K}^{(l)}\left(z^{(l)}\right), \quad z^{(l+1)} = \tilde z^{(l)} + \alpha A^{(l)} \tilde z^{(l)}
\end{equation}
This structure provides stable geometric constraints for latent space evolution without altering the expressive power of the original operators.

\subsection{Theoretical Analysis} 
\pb{Stability Analysis.}
The neural operator inter-layer update is performed through the following.
\begin{equation}
    z^{+} = (I + \alpha A) z,
\end{equation}
where $A^\top = -A$. 
Rewriting the equation, we have as follows. 
\begin{equation}
    \|z^{+}\|_2^2 = z^\top (I + \alpha A)^\top (I + \alpha A) z = \|z\|_2^2 + \alpha^2 \|A z\|_2^2
\end{equation}
Note that the change in the norm is only a second-order term, indicating that this update is numerically stable under the condition of small step sizes.

\pb{Geometric Consistency.}
The update direction is always within the tangent space of the manifold.
Thus, the evolution of the hidden state is restricted to the neighborhood of isometric transformation, thereby effectively suppressing the diffusion of features along non-physical directions. 
This property is particularly important in multi-layer stacking and long-term inference. 

\pb{Complexity Analysis.}
Given that we utilize a low-rank approximation of dimension $r$ to construct the generator of latent space in of channel $C$ through $N$ sampling, our method only imposes $O(Cr)$ parameter complexity.
This is executed within $O(CrN)$ time and consumes only $O(Cr)$ memory space.
This is because $r \ll C$ creates negligible overhead compared to the standard neural operator.
We provide more theoretical analysis in Appendix~\ref{app:theory}.
\section{Experiments}
\subsection{Baseline Models and Datasets}
To evaluate our method, we leverage three widely implemented neural operators: Fourier Neural Operator (FNO)~\cite{li2020fourier}, U-shaped Neural Operator (U-NO)~\cite{rahman2022u}, and Geometry-Informed Neural Operator (GINO)~\cite{li2020geometry}.
We integrate MCL in every spectral layer of these models. 
Specifically, we insert MCL after the activation function in FNO and between spectral convolution and activation function for U-NO, and GINO.
We provide more information on implementation details on these neural operators in Appendix~\ref{app:model}.
We evaluate our method on several fundamental PDEs in fluid mechanics.
Specifically, we construct the data for model training and evaluation by strictly following the physical conservation laws and a high-precision numerical solution w.r.t each unique equation characteristic.
For instance, we combine the pseudo-spectral method with Crank-Nicolson or Runge-Kutta time integration to ensure the physical consistency of trajectories. 
For the one-dimensional benchmark test, we select the Burgers equation~\cite{bonkile2018systematic}, the linear advection equation~\cite{mojtabi2015one}, and the diffusion-sorption equation
~\cite{takamoto2022pdebench}.
Burgers equation combines nonlinear advection and diffusion terms and is often regarded as a microcosm of the Navier-Stokes equation. 
For the two-dimensional benchmark test, we utilize the elliptic Darcy flow equation~\cite{hubbert1956darcy}, the hyperbolic shallow water equation~\cite{takamoto2022pdebench}, and the Navier-Stokes equation~\cite{constantin1988navier}.
The former describes the pressure distribution in heterogeneous permeability fields in porous media, while the latter characterizes the strong nonlinear coupling mechanism of free surface fluids. 
We generate these test sets by mixing various physical parameters (e.g., viscosity, convection velocity, and permeability) to comprehensively verify the generalization ability and robustness of the model when dealing with shock waves, wave propagation, and complex parameter field variations. 
We provide more explanation on data construction in Appendix~\ref{app:data}.

\begin{table*}[!htb]
    \small
    \centering
    \caption{Evaluation on MCL integration to neural operator models under various equations, parameters, and rollout steps. Best results are in \textbf{bold}.}
    \label{tab:main}
    \begin{tabular}{p{50pt}ccccccccc}
        \toprule
        \multirow{2}*{Equation} & \multirow{2}*{Param.} & \multirow{2}*{Step} & \multirow{2}*{Model} &\multicolumn{3}{c}{Model only} & \multicolumn{3}{c}{Model + MCL}\\
        \cmidrule(lr){5-7} \cmidrule(lr){8-10}
        &&&& MSE & Rel\_L2 & Rel\_H1 & MSE & Rel\_L2 & Rel\_H1\\
        \midrule
         \multirow{6}{50pt}{1-D Burgers} & \multirow{6}*{0.001} & \multirow{3}*{1} & FNO & 0.000550 & 0.046626 & 0.068925 & 0.000327 & \textbf{0.035689} & 0.049867\\
         &&& U-NO & 0.042886 & 0.398778 & 0.415302 & 0.039507 & \textbf{0.350689} & 0.367253\\
         &&& GINO & 0.003950 & 0.147548 & 0.192726 & 0.003891 & \textbf{0.146621} & 0.192288\\
         \cmidrule(lr){3-10}

         && \multirow{3}*{10} & FNO & 0.000214 & 0.037893 & 0.050876 & 0.000108 & \textbf{0.026429} & 0.034554\\
         &&& U-NO & 0.025308 & 0.412722 & 0.417508 & 0.020423 & \textbf{0.398555} & 0.404843\\
         &&& GINO & 0.001475 & 0.109261 & 0.139855 & 0.001446 & \textbf{0.107881} & 0.138665\\
         \cmidrule(lr){3-10}
         
         && \multirow{3}*{25} & FNO & 0.000098 & 0.032601 & 0.039601 & 0.000046 & \textbf{0.022443} & 0.026869\\
         &&& U-NO & 0.014334 & 0.373097 & 0.376473 & 0.008402 & \textbf{0.331120} & 0.334335\\
         &&& GINO & 0.000586 & 0.082725 & 0.102302 & 0.000573 & \textbf{0.081434} & 0.101060\\
         \midrule

         \multirow{6}{50pt}{1-D Advection} & \multirow{6}*{1.0} & \multirow{3}*{1} & FNO & 0.000243 & 0.013964 & 0.015500 & 0.000064 & \textbf{0.008839} & 0.009975\\
         &&& U-NO & 0.013008 & 0.189304 & 0.195135 & 0.012668 & \textbf{0.174245} & 0.178913\\
         &&& GINO & 0.003759 & 0.074602 & 0.089012 & 0.001234 & \textbf{0.038764} & 0.045873\\
         \cmidrule(lr){3-10}
         && \multirow{3}*{25} & FNO & 0.000217 & 0.014444 & 0.016174 & 0.000060 & \textbf{0.009033} & 0.010409\\
         &&& U-NO & 0.042698 & 0.289166 & 0.290737 & 0.019341 & \textbf{0.187799} & 0.189771\\
         &&& GINO & 0.004569 & 0.081369 & 0.095214 & 0.001297 & \textbf{0.039893} & 0.046892\\
         \midrule

         \multirow{6}{50pt}{1-D Diffusion} & \multirow{6}*{-} & \multirow{3}*{1} & FNO & 0.000013 & 0.013330 & 0.013673 & 0.000000 & \textbf{0.001321} & 0.002699\\
         &&& U-NO & 0.016053 & 0.444866 & 0.447889 & 0.005528 & \textbf{0.264650} & 0.269736\\
         &&& GINO & 0.000138 & 0.043518 &  0.048344 & 0.269736 & \textbf{0.024395} & 0.025905\\
         \cmidrule(lr){3-10}
         && \multirow{3}*{25} & FNO & 0.000001 & 0.001751 & 0.002732 & 0.000000 & \textbf{0.001016} & 0.002226\\
         &&& U-NO & 0.015116 & 0.346884 & 0.347008 & 0.007619 & \textbf{0.235791} & 0.236084\\
         &&&  GINO &0.000028 & 0.014957 & 0.017041& 0.000005 & \textbf{0.005968} & 0.007054\\
         \midrule

         \multirow{6}{50pt}{2-D Darcy Flow} & \multirow{3}*{0.01} & \multirow{3}*{1} & FNO & 0.000572 & 2.596873 & 2.617427 & 0.000194 &\textbf{ 1.040141} & 1.063479\\
         &&& U-NO & 0.000228 & 1.692483 & 1.784757 & 0.000211 & \textbf{1.362368} & 1.647808\\
         &&& GINO & 0.001068 & 5.074404 & 5.058982 & 0.001068 & \textbf{5.068308} & 5.052908\\
         \cmidrule(lr){3-10}
         
         & \multirow{3}*{1.0} & \multirow{3}*{1} & FNO & 0.000857 & 0.124781 & 0.128276 & 0.000307 & \textbf{0.054643} & 0.060323\\
         &&& U-NO & 0.000360 & 0.068732 & 0.076765 & 0.000268 & \textbf{0.024928} & 0.026717\\
         &&& GINO & 0.018728 & 0.672266 & 0.675335 & 0.002421 & \textbf{0.244659} & 0.283356\\
         \midrule
         
         \multirow{6}{50pt}{2-D Shallow Water} & \multirow{6}*{-} & \multirow{3}*{1} & FNO & 0.000021 & 0.004370 & 0.008327 & 0.000010 & \textbf{0.003084} & 0.005210 \\
         &&& U-NO & 0.035991 & 0.172072 & 0.172072 & 0.012362 & \textbf{0.101253} & 0.112789\\
         &&& GINO & 0.020287 & 0.131313 & 0.149931 & 0.020275 & \textbf{0.131233} & 0.150866\\
         \cmidrule(lr){3-10}
         && \multirow{3}*{25} & FNO & 0.000010 & 0.003055 & 0.004681 & 0.000003 & \textbf{0.001657} & 0.002351\\
         &&& U-NO & 0.020450 & 0.136177 & 0.138663 & 0.009696 & \textbf{0.089772} & 0.092683\\
         &&& GINO & 0.008818 & 0.086397 & 0.098830 & 0.008813 & \textbf{0.086314} & 0.098691\\
         \midrule

         \multirow{6}{50pt}{2-D Navier-Stokes} & \multirow{6}*{-} & \multirow{3}*{1} & FNO & 0.094958 & 0.006878 & 0.007343 & 0.003067 & \textbf{0.005678} & 0.006499\\
         &&& U-NO & 0.121584 & 0.090440 & 0.099027 & 0.117830 & \textbf{0.085402} & 0.094537\\
         &&& GINO & 0.014434 & 0.013638 & 0.021241 & 0.013268 & \textbf{0.013185} & 0.020583\\
         \cmidrule(lr){3-10}

         &&\multirow{3}*{5} & FNO & 0.002618 & 0.006557 & 0.007242 & 0.002553 & \textbf{0.005167} & 0.006075\\
         &&& U-NO & 0.062031 & 0.044698 & 0.052281 & 0.044285 & \textbf{0.039946} & 0.046540\\
         &&& GINO & 0.011598 & 0.011943 & 0.019204 & 0.010699 & \textbf{0.011584} & 0.018829\\
         \cmidrule(lr){3-10}
         
         && \multirow{3}*{10} & FNO & 0.003486 & 0.007163 & 0.008197 & 0.003328 & \textbf{0.005932} & 0.007190\\
         &&& U-NO & 0.064937 & 0.059068 & 0.064250 & 0.051550 & \textbf{0.051766} & 0.057728\\
         &&& GINO & 0.014704 & 0.013079 & 0.019682 & 0.012966 & \textbf{0.012962} & 0.019650\\
        \bottomrule
    \end{tabular}
\end{table*}

\subsection{Experimental Setup}
We conduct all experiments on an NVIDIA RTX 4090 24G using the PyTorch framework. 
We set the training data size for each parameter to 1000, while the test data consisted of 200 independently generated trajectories. 
We leverage MSE loss as a training objective to evaluate prediction accuracy.
We utilize Adam optimizer~\cite{kingma2014adam} with an initial learning rate set to 0.001 with a cosine annealing scheduler~\cite{loshchilov2017sgdr}.
We train our model for 500 epochs with the pushforward method with $T=5$.
To examine the model's generalization to parameter variations, we mix various uniformly samples parameter values to generate training data for mixed parameter training on a single model.
We generate separate validation datasets for each specific parameter value to perform independent evaluations for verifying the model's robustness and generalization performance. 
To ensure resolution invariance, we train with data downsampled to a fixed resolution (e.g., 256 for 1-D equations and $64 \times 64$ for 2-D ones). 
During inference, we perform a resampling scheme by downsampling the data to the resolution used during training, and upsampling it back to the original resolution after processing is finished. 
This is to preserve fine-grained information while inheriting the zero-shot super-resolution capability of the FNO.
We set the low-rank dimension to $r=8$, and step size $\alpha$ is left as learnable.

\subsection{Evaluation Metrics}
We adopt a multi-dimensional evaluation system to comprehensively evaluate the model's performance in numerical accuracy, physical conservation, and local extremum capture. 
In addition to the standard relative error, we introduce an absolute error metric to measure the actual deviation of physical quantities.
All metrics are calculated as averages on the test set. 

\pb{Relative L2 Error.}
The relative L2 error (Rel\_L2) evaluates the proportion of the energy belong to the error to the total energy of the true signal. 
This metric is defined as follows.
\begin{equation}
    \text{Rel\_L2} = \frac{\|u - \hat{u}\|_2}{\|u\|_2} = \frac{\sqrt{\sum (u_i - \hat{u}_i)^2}}{\sqrt{\sum u_i^2}}
\end{equation}
We implement the following gold standard in the field of machine learning for science to evaluate the numerical precision of the results:
\begin{itemize}
    \item $< 1\% (10^{-2})$: high precision, approaching conventional numerical method
    \item $< 10\% (10^{-1}$: acceptable baseline performance
    \item $> 1.0$: prediction failure
\end{itemize}

\pb{Relative H1 Error.}
The relative H1 (Sobolev) error (Rel\_H1) measures the relative error of the function value and its gradient.
This metric complement relative $L2$ error by examining the function shape instead of solely relying on the predicted value.
This metric is defined as follows.
\begin{equation}
\text{Rel\_H1} = \frac{\|u - \hat{u}\|_{H^1}}{\|u\|_{H^1}} = \sqrt{\frac{\|u - \hat{u}\|_2^2 + \alpha \|\nabla u - \nabla \hat{u}\|_2^2}{\|u\|_2^2 + \alpha \|\nabla u\|_2^2}}
\end{equation}
For equations that are sensitive to derivatives, such as the Kuramoto-Sivashinsky equation, the relative $L2$ error may be very low, but the direction of the derivatives may not adhere to the true function.
In this metric, a low relative $H1$ error indicates that the model has nearly identical shape and direction of derivatives to the true function, which is necessary in ensuring the long-term prediction stability.

\pb{Mean Squared Error.}
The mean squared error (MSE) measures the average of the squares of the differences between the predicted values and the true values.
This metric is defined as follows.
\begin{equation}
    \text{MSE} = \frac{1}{N} \sum_{i=1}^{N} (u_i - \hat{u}_i)^2
\end{equation}

\pb{Rolling Deduction Indicator.}
We leverage a rolling (autoregressive) mechanism to evaluate the long-term behavior of the model.
Specifically, we feed the prediction of the previous time step as input to the next time step to allow for error accumulation.
This evaluation protocol is designed to assess the prediction stability, physical consistency, and long-term dynamical fidelity of the model.
To quantify how errors propagate during rollout, we report the rollout MSE (rollout\_MSE) and rollout relative L2\textsuperscript{T} error (rollout\_rel\_L2\textsuperscript{T}) over the entire prediction steps.
We also measure rollout relative H1 error (rollout\_rel\_H1) at selected rollout steps.

\begin{figure}[!htb]
    \centering
    \includegraphics[width=\linewidth]{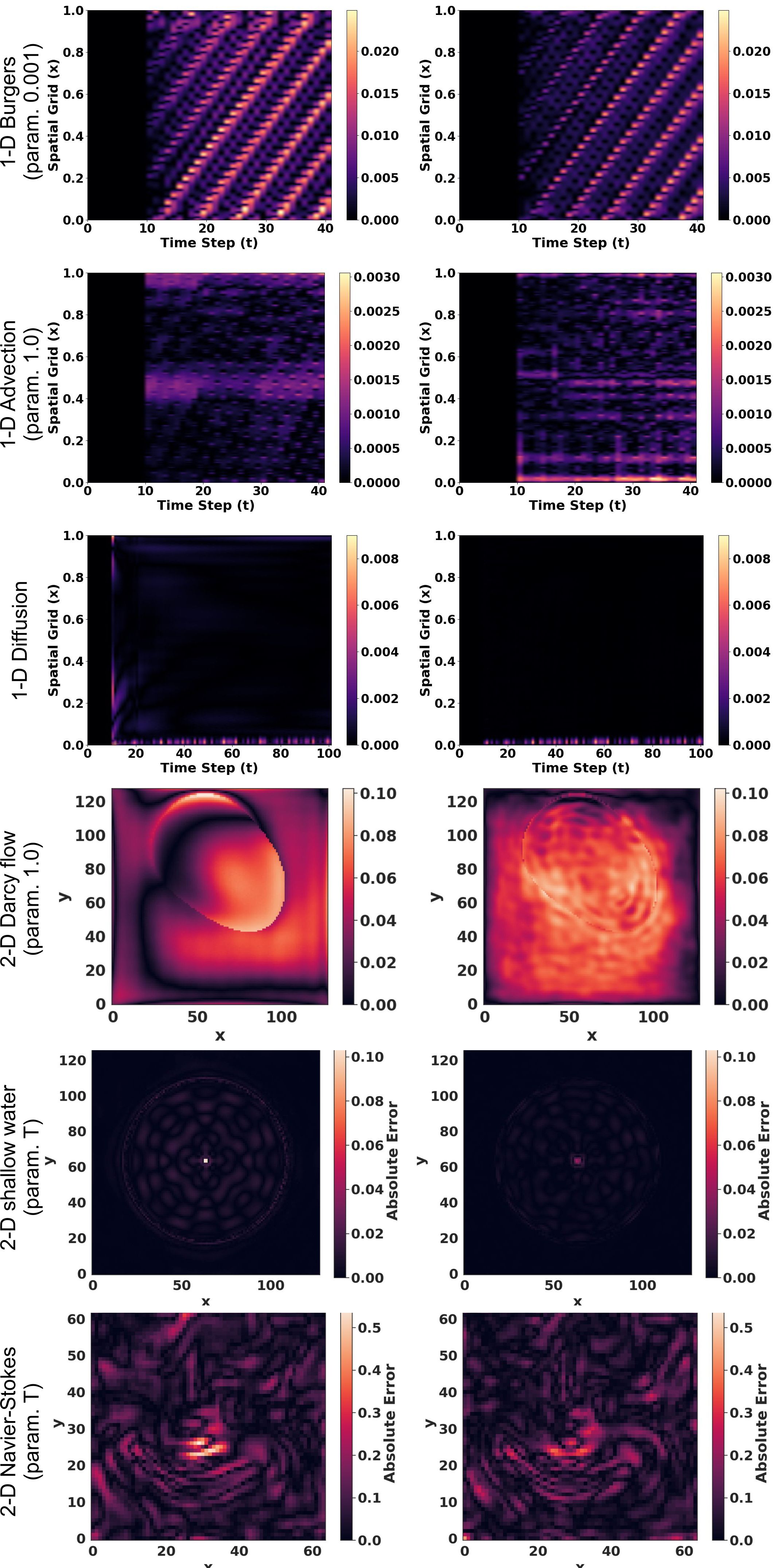}
    \caption{Prediction error plot of PDEs on FNO (left) and FNO+MCL (right) against true solution.}
    \label{fig:prediction}
\end{figure}

\subsection{Main Results}
We present the result of the main experiment in Table~\ref{tab:main}.
Across nearly all equations, incorporating MCL to all neural operator models consistently reduces MSE, relative L2 error, and relative H1 error.
Moreover the relative errors experience the most pronounced improvement, indicating that MCL primarily improves the global accuracy and the fitness to the true function shape and derivative direction.
Importantly, the gains from MCL persist and often amplify at longer rollout steps.
This suggests that MCL mitigates error accumulation and improves the long-term stability of model dynamics.
We provide more experiment results on various parameters and step choices of all PDEs in Appendix~\ref{app:results}.

\pb{Prediction Error Plot.}
As shown in Figure~\ref{fig:prediction}, MCL consistently improves the preservation of spatiotemporal structure compared to the FNO. 
For 1-D Burgers’ and linear advection (first and second rows), MCL maintains sharper and better-aligned diagonal wave patterns over time, while FNO exhibits noticeable phase drift and amplitude attenuation. 
In the 1-D diffusion-sorption case (third row), both models capture the diffusive trend with MCL produces smoother and more physically consistent gradients during long-term evolution. 
For 2-D Darcy flow and shallow water equations (fourth and fifth rows), MCL better preserves global spatial structure and symmetry, avoiding the over-smoothing observed in the FNO. 
Finally, in 2-D Navier-Stokes (bottom row), MCL retains more coherent vortical patterns with reduced artificial diffusion. 
Overall, the prediction plot demonstrates that Lie algebra-parameterized manifold evolution mitigates phase drift and structural degradation during rollout, leading to more stable and physically consistent operator learning across diverse PDEs.

\begin{figure}[!htb]
    \centering
    \includegraphics[width=.9\linewidth]{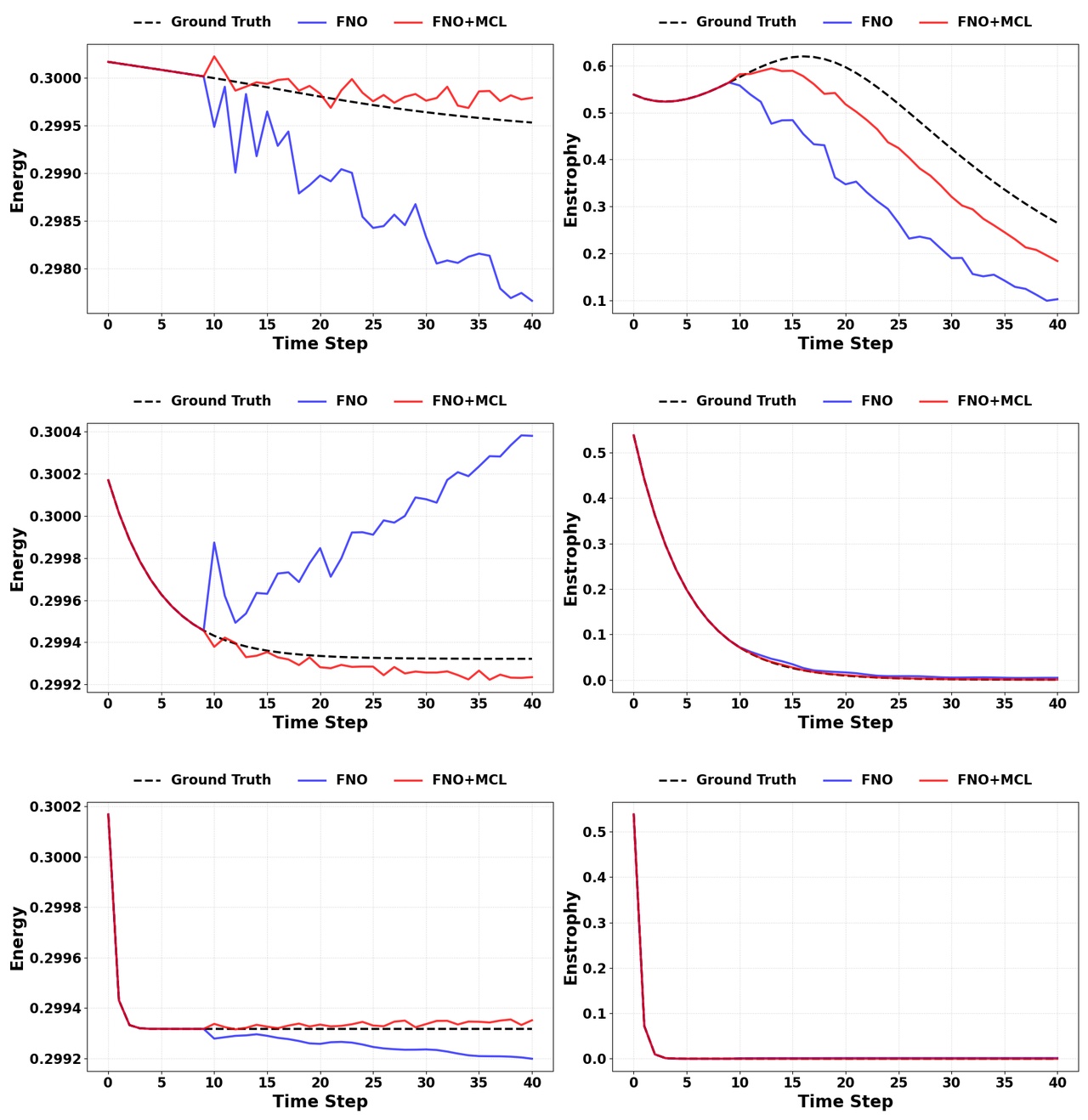}
    \caption{Evolution of energy (left) and entropy (right) on 1-D Burgers equation with parameter values 0.001 (top), 0.01 (middle), and 0.1 (bottom).}
    \label{fig:latent}
\end{figure}

\pb{Latent Drift Reduction Analysis.}
To evaluate the effectiveness of MCL on addressing latent drift, we visualize the evolution of energy and entropy on Figure~\ref{fig:latent}.
The energy plot shows that FNO suffers from a gradual upward drift during rollout, which increases deviation from the true solution.
In contrast, the integration of MCL closely follows the true solution.
The entropy plot further supports these observations.
In 1-D Burgers equation, all methods exhibit decay due to viscous dissipation.
Nevertheless, MCL integration remains consistently aligned with the true solution, even at later time steps.
Meanwhile, FNO retains slightly higher residual entropy.
These results demonstrate that enforcing manifold constraints on representation update via Lie algebra parameterization effectively alleviates latent drift and improves long-horizon stability.

\subsection{Ablation Studies}

\pb{Contribution of MCL.}
We evaluate the contribution of MCL in improving the performance of neural operator model by replacing it with a standard MLP augmentation.
Table~\ref{tab:contribution} demonstrates that across all parameter values and rollout steps, integrating MCL consistently outperforms MLP augmentation in evaluation metrics. 
The performance gap between MCL and MLP widens at step 25 in the relative error metrics. 
This indicates that MCL is more effective in suppressing error accumulation during rollout, which leads to more stable long-term predictions.

\begin{table}[!htb]
    \small
    \centering
    \caption{Ablation on MCL contribution on 1-D Advection. Best results are in \textbf{bold}.}
    \label{tab:contribution}
   \begin{tabular}{p{20pt}ccccc}
        \toprule
        Param. & Step & Model & MSE & Rel\_L2 & Rel\_H1\\
        \midrule

        \multirow{4}{*}{0.1} & \multirow{2}{*}{1} & FNO\textsubscript{+MCL} & \textbf{0.000016} & \textbf{0.004433} & \textbf{0.005201}\\ 
        && FNO\textsubscript{+MLP} & 0.000019 & 0.004549 & 0.005406\\
        \cmidrule{2-6}
         & \multirow{2}{*}{25} & FNO\textsubscript{+MCL} & \textbf{0.000031} & \textbf{0.006246} & \textbf{0.006954}\\ 
         && FNO\textsubscript{+MLP} & 0.000034 & 0.008008 & 0.008762\\
         \midrule
         \multirow{4}{*}{1.0} & \multirow{2}{*}{1} & FNO\textsubscript{+MCL} & \textbf{0.000064} & \textbf{0.008839} & \textbf{0.009975}\\ 
         && FNO\textsubscript{+MLP} & 0.000078 & 0.009654 & 0.010835\\
         \cmidrule{2-6}
         & \multirow{2}{*}{25} & FNO\textsubscript{+MCL} & \textbf{0.000060} & \textbf{0.009033} & \textbf{0.010409}\\ 
         && FNO\textsubscript{+MLP} & 0.000074 & 0.009923 & 0.011380\\
        \bottomrule
         
    \end{tabular}
\end{table}

\pb{Model Parameter.}
We measure the number of parameter affected by the integration of MCL to the neural operator models.
Table~\ref{tab:param} shows that the addition of parameters from MCL to the existing neural operators is significantly less than the parameters of the model itself.
On average, MCL only increases the model parameter by 2.26\%.
This demonstrates that MCL improves the performance of neural operator at a minimum overhead.

\begin{table}[!htb]
    \small
    \centering
    \caption{Total parameter count of original model vs. integration with MCL.}
    \label{tab:param}
   \begin{tabular}{cccc}
        \toprule
        Model & \# of param. & \# of param+MCL & \% param. add\\
        \midrule

        FNO & 23,937 & 25,221 & 5.36\\ 
        U-NO & 2,708,673 & 2,732,234 & 0.87\\
        GINO & 442,194 & 444,567 & 0.54\\
        \bottomrule
         
    \end{tabular}
\end{table}

\pb{Hyperparameter Sensitivity.}
We evaluate the model sensitivity to the choice of rank dimension.
Table~\ref{tab:sensitivity} shows that rank 3 and lower consistently show higher relative errors.
Meanwhile, medium rank (e.g., 8-10) yields the lower relative errors are both short and medium steps, supporting our chosen rank dimension at 8.
Rank higher than 10, on the other hand, gives only negligible improvements.
This result shows that as rank increases, the model performance continue to improve until it is saturated at rank beyond 10.

\begin{table}[!htb]
    \small
    \centering
    \caption{Sensitivity test on hyperparameter.}
    \label{tab:sensitivity}
   \begin{tabular}{ccccc}
        \toprule
        Rank dim. & Step & MSE & Rel\_L2 & Rel\_H1\\
        \midrule
        \multirow{4}{*}{3} & 1 & 0.000021 & 0.000020 & 0.000017\\ 
        & 10 & 0.000003 & 0.004529 & 0.004668\\
        & 25 & 0.000001 & 0.003394 & 0.003454\\
        & 30 & 0.000001 & 0.003382 & 0.003436\\
        \midrule
        \multirow{4}{*}{5} & 1 & 0.000020 & 0.008317 & 0.008839\\ 
        & 10 & 0.000003 & 0.004339 & 0.004482\\
        & 25 & 0.000001 & 0.003393 & 0.003449\\
        & 30 & 0.000001 & 0.003339 & 0.003389\\
        \midrule
        \multirow{4}{*}{8} & 1 & 0.000017 & \textbf{0.007182} & \textbf{0.007599}\\ 
        & 10 & 0.000003 & \textbf{0.004170} & \textbf{0.004294}\\
        & 25 & 0.000001 & 0.003402 & 0.003457\\
        & 30 & 0.000001 & 0.003255 & \textbf{0.003307}\\
        \midrule
        \multirow{4}{*}{10} & 1 & 0.000018 & 0.007460 & 0.007927\\ 
        & 10 & 0.000003 & 0.004410 & 0.004543\\
        & 25 & 0.000001 & \textbf{0.003375} & \textbf{0.003426}\\
        & 30 & 0.000001 & 0.003304 & 0.003350\\
        \midrule
        \multirow{4}{*}{12} & 1 & 0.000018 & 0.007583 & 0.008033\\ 
        & 10 & 0.000003 & 0.004286 & 0.008033\\
        & 25 & 0.000001 & 0.003416 & 0.003474\\
        & 30 & 0.000001 & \textbf{0.003124} & 0.003177\\
        \bottomrule
    \end{tabular}
\end{table}

\section{Conclusion}
Neural operators provide a powerful paradigm for learning solutions to partial differential equations.
Its expressiveness is, however, limited by feature evolution instability under iterative and long-horizon prediction.
We address this limitation by enforcing manifold-constrained latent evolution through Lie algebra parameterization.
This geometric constraining effectively mitigates latent drift while remaining lightweight and applicable to various existing neural operators.
Our work suggests a promising direction for geometry-aware neural operator design.

\section{Impact Statement}  
Our method improves the stability and reliability of neural operators for modeling physical systems.
This enables more accurate long-term predictions with minimal additional computational cost. 
By enforcing geometry-aware latent evolution, our method reduces error accumulation and numerical divergence, which are critical barriers to deploying neural operators in real-world scientific and engineering applications. 
Finally, our work supports safer and more trustworthy data-driven simulation tools for domains such as climate modeling, fluid dynamics, and engineering design.

\bibliography{example_paper}
\bibliographystyle{icml2026}

\newpage
\appendix
\onecolumn
\setcounter{table}{0}
\renewcommand{\thetable}{A\arabic{table}}
\setcounter{figure}{0}
\renewcommand{\thefigure}{A\arabic{figure}}
\setcounter{section}{0}
\renewcommand{\thesection}{A}
\subsection{Data Construction}\label{app:data}
\pb{1-D Burgers Equation.}
The Burgers equation~\cite{bonkile2018systematic} is the most simplified model for studying nonlinear acoustics and shock wave dynamics in fluid mechanics. 
It is often regarded as a one-dimensional microcosm of the Navier-Stokes equation. 
The Burgers equation is defined as follows.
\begin{equation}
    \frac{\partial u}{\partial t} + u \frac{\partial u}{\partial x} = \nu \frac{\partial^2 u}{\partial x^2}, \quad x \in [0, 2\pi)
\end{equation}
The left term contains the nonlinear advection associated with conservation laws ($u u_x$), while the right part contains the diffusion associated with heat equation ($\nu u_{xx}$).
Notice that this equation requires solving the derivatives both in the spatial and temporal domains.
To compute the spatial derivatives, we adopt a pseudo-spectral method that expands the spatial solution to spectral basis functions and computes the nonlinear term in the physical space.
To compute temporal derivatives, we utilize Adams-Bashforth equation for the nonlinear advection terms and Crank-Nicolson equation for the linear diffusion term~\cite{he2007stability}.
We set the physical parameter $\nu \in \{0.1, 0.01, 0.001\}$ and discretization at $N_x=1024$ and $t \in [0, 1.0]$.

\pb{1-D Linear Advection Equation.} 
The linear advection equation~\cite{mojtabi2015one} describes wave propagation or material transport phenomena and is commonly used to test the phase shift error and numerical dissipation of numerical schemes. 
As the inviscid limit of the Burgers equation, it can examine the model's ability to handle pure convective dynamics.
The linear advection equation is expressed as follows.
\begin{equation}
    \frac{\partial u}{\partial t} + \beta \frac{\partial u}{\partial x} = 0, \quad x \in [0, 2\pi)
\end{equation}
We leverage the pseudo-spectral method~\cite{fornberg1998practical} to compute the spatial derivatives.
Meanwhile, we compute the temporal derivatives with Runge-Kutta equation~\cite{butcher2007runge}.
We set the constant velocity term $\beta \in \{0.1, 1.0, 4.0\}$ and discretization at $N_x=1024$ and $t \in [0, 1.0]$, discretized into 100 time steps.

\pb{1-D Diffusion-Sorption Equation.}
The diffusion process~\cite{takamoto2022pdebench} delayed by the adsorption process is widely used in real-world scenarios such as groundwater pollutant transport.
This process describes the exchange of solutes between the solid and liquid phases through a non-linear retardation factor $R(u)$.
The diffusion-sorption equation is expressed as follows.
\begin{equation}
    \frac{\partial u}{\partial t} = \frac{D}{R(u)} \frac{\partial^2 u}{\partial x^2}, \quad x \in (0, 1), \quad t \in (0, 500]
\end{equation}
The retardation factor follows the Freundlich adsorption isotherm~\cite{appel1973freundlich}.
\begin{equation}
    R(u) = 1 + \frac{1-\phi}{\phi} \rho_s k n_f u^{n_f-1}
\end{equation}
To solve the spatial and temporal derivatives of the diffusion-sorption equation, we utilize the finite volume method (FVM)~\cite{eymard2000finite} and the fourth-order Runge-Kutta~\cite{carpenter2005fourth}, respectively.
We set the constant terms for porosity $\phi=0.29$, density $\rho_s=2880$, adsorption coefficient $k=3.5 \times 10^{-4}$, exponent $n_f=0.874$, and diffusion coefficient $D=5 \times 10^{-4}$. 
We perform discretization at a spatial resolution of $N_s$ = 1024 and time domain sampling for 100 steps.
We directly train this dataset without downsampling.

\pb{2-D Darcy Flow Equation.}
The Darcy flow equation~\cite{hubbert1956darcy} describes fluid flow in porous media and is commonly used in groundwater simulation and oil extraction. 
As an elliptic partial differential equation, it involves the calculation of pressure distribution and velocity under non-uniform permeability fields and tests the robustness of the model against variations in the parameter field.
The Darcy flow equation is expressed as follows.
\begin{equation}
    -\nabla \cdot (\beta \nabla p) = f, \quad (x,y) \in [0,1] \times [0,1]
\end{equation}
We solve this equation using finite element method~\cite{reddy1993introduction} with the source term $f$ set as random Gaussian field.
We set the constant term for permeability $\beta \in \{0.01, 0.1, 1.0\}$ and perform discretization in spatial dimension $128 \times 128$.

\pb{2-D Shallow Water Equation.}
Shallow water equation~\cite{takamoto2022pdebench} is a hyperbolic partial differential equation that describes the flow of a fluid with a free surface under the action of gravity.
This equation couples several nonlinear parameters, such as fluid height $h$, horizontal velocity $u$, and vertical velocity $v$.
The shallow water equation is defined as follows.
\begin{equation}
    \begin{cases} h_t + \nabla \cdot (h \mathbf{u}) = 0 \\ \mathbf{u}_t + (\mathbf{u} \cdot \nabla)\mathbf{u} + g \nabla h = 0 \end{cases}
\end{equation}
We construct the spatial and nonlinear terms of these equation in a rotational form.
Specifically, we solve the temporal derivatives using the fourth-order Runge-Kutta and apply $\frac{2}{3}$ dealiasing rule. 
We set the constant terms for gravitational acceleration $g=1.0$, average water depth $H_0=1.0$. 
We utilize a random Gaussian perturbation as the initial condition with the perturbation radius set at $r \in (0.3, 0.7)$.
We perform discretization for spatial resolution $64 \times 64$ and temporal domain $T=2.0$. 

\pb{2-D Navier-Stokes Equations.}
Navier-Stokes equations~\cite{constantin1988navier} describe compressible fluid dynamics.
These are capable of simulating the formation and propagation of shock waves and turbulent dynamics. 
The equations are utilized to study complex real-world problems such as aerodynamics (e.g., aircraft wings) and astrophysics (e.g., interstellar gas dynamics).
Navier-Stokes equations are a system of partial differential equation comprising several aspects, including mass conservation, momentum conservation, and energy conservation.
The equation system is describes as follows.
\begin{equation}
\begin{aligned}
    \partial_t \rho + \nabla \cdot (\rho v) &= 0\\
    \rho(\partial_t v + v \cdot \nabla v) &= -\nabla p + \eta \triangle v \\
    &+ (\zeta + \eta/3)\nabla(\nabla \cdot v)\\
\end{aligned}
\end{equation}
We solve the inviscid terms of the equations with the second-order Harten-Lax-van Leer-Contact scheme~\cite{harten1983upstream}, combined with the monotonic upstream-centered schemes~\cite{van1977towards} for conservation laws.
We set the constant terms $M = 0.1$ and $\zeta = 0.01$.
The discretization is performed ar spatial resolution $512 \times 512$ with 21 time steps to generate 1k samples.
We utilize this dataset for training with downsampling at $256 \times 256$ resolution.

\subsection{Theoretical Analysis: Lie-Algebraic Near-Isometric Hidden Dynamics}
\label{app:theory}

This section provides a rigorous analysis of the numerical stability and geometric consistency of the proposed Manifold-Constrained Layer (MCL) from the perspective of hidden-state dynamics and structure-preserving updates. The central result is that MCL replaces unconstrained Euclidean residual updates with Lie-algebra-generated near-isometric evolution, under which first-order norm drift vanishes identically and hidden-state variations are dominated by second-order terms. This mechanism effectively suppresses latent drift in deep stacking and long-horizon rollout settings.

\pb{Hidden-State Updates as Discrete Dynamical Systems.}
Standard neural operators update the hidden representation at layer $l$ according to
\begin{equation}
z^{(l+1)} = z^{(l)} + \mathcal{K}^{(l)}\!\left(z^{(l)}\right), 
\qquad z^{(l)} \in \mathbb{R}^{C \times N},
\label{eq:euclidean_update}
\end{equation}
where $C$ denotes the channel dimension, $N$ the number of spatial discretization points, and $\mathcal{K}^{(l)}(\cdot)$ the operator mapping at layer $l$ (e.g., spectral or local convolution). This update can be interpreted as an explicit Euler discretization of an underlying hidden-state dynamical system in Euclidean space. When repeatedly composed across layers or rollout steps, such unconstrained updates may lead to progressive deviation of hidden states from their effective low-dimensional structure, resulting in norm growth and accumulated numerical instability.

\pb{Lie Group-Constrained Near-Isometric Updates.}
To impose structural constraints on hidden-state evolution, MCL assumes that layerwise updates are generated by Lie group actions:
\begin{equation}
z^{(l+1)} = G^{(l)} z^{(l)}, 
\qquad G^{(l)} \in \mathfrak{so}(C),
\label{eq:group_action}
\end{equation}
where $\mathfrak{so}(C)$ denotes the orthogonal group satisfying
\begin{equation}
\big(G^{(l)}\big)^\top G^{(l)} = I .
\label{eq:orthogonality}
\end{equation}
Such transformations preserve inner products and correspond to isometries in the channel space.
Rather than optimizing $G^{(l)}$ directly, MCL employs Lie algebra parameterization. Let
\begin{equation}
A^{(l)} \in \mathfrak{so}(C), 
\qquad (A^{(l)})^\top = -A^{(l)},
\label{eq:lie_algebra}
\end{equation}
and define the group element via the exponential map
\begin{equation}
G^{(l)} = \exp\!\big(A^{(l)}\big).
\label{eq:exp_map}
\end{equation}
Since the exponential of a skew-symmetric matrix is orthogonal, the continuous update preserves the hidden-state norm exactly.
For computational efficiency, a first-order approximation of the exponential map is adopted. For a sufficiently small step size $\alpha>0$,
\begin{equation}
\exp\!\big(\alpha A^{(l)}\big) \approx I + \alpha A^{(l)},
\label{eq:linearized_exp}
\end{equation}
which yields the practical MCL update
\begin{equation}
z^{(l+1)} = z^{(l)} + \alpha A^{(l)} z^{(l)}.
\label{eq:mcl_update}
\end{equation}
This update advances hidden states along directions in the tangent space of the manifold induced by $so(C)$, rather than introducing arbitrary Euclidean residual perturbations.

\pb{Low-Rank Lie Algebra Parameterization.}
To reduce computational and memory costs, MCL adopts a low-rank skew-symmetric construction:
\begin{equation}
A^{(l)} = U^{(l)}\big(V^{(l)}\big)^\top - V^{(l)}\big(U^{(l)}\big)^\top,
\qquad
U^{(l)}, V^{(l)} \in \mathbb{R}^{C \times r}, \quad r \ll C.
\label{eq:low_rank_A}
\end{equation}
This form satisfies $(A^{(l)})^\top = -A^{(l)}$ by construction. Its action on a vector $z$ can be expanded as
\begin{equation}
A^{(l)} z
= U^{(l)}\!\left(\big(V^{(l)}\big)^\top z\right)
- V^{(l)}\!\left(\big(U^{(l)}\big)^\top z\right),
\label{eq:low_rank_action}
\end{equation}
which avoids explicit $C\times C$ matrix construction. The resulting computational complexity is $O(Cr)$ per spatial location and $O(CrN)$ overall, which is negligible when $r \ll C$.

\pb{Integration with Neural Operator Layers.}
MCL is integrated with a generic operator layer $K^{(l)}$ in a serial manner:
\begin{equation}
\tilde z^{(l)} = \mathcal{K}^{(l)}\!\big(z^{(l)}\big), 
\qquad
z^{(l+1)} = \tilde z^{(l)} + \alpha A^{(l)} \tilde z^{(l)}.
\label{eq:serial_integration}
\end{equation}
In this formulation, the operator layer governs representational mapping, while MCL constrains the subsequent hidden-state evolution to remain within a near-isometric neighborhood.

\pb{Single-Step Norm Stability.}
Consider the single-step update
\begin{equation}
x^{+} = (I + \alpha A) x,
\qquad A^\top = -A, \quad x \in \mathbb{R}^C .
\label{eq:single_step}
\end{equation}
Its squared $\ell_2$ norm satisfies
\begin{align}
\|x^{+}\|_2^2
&= x^\top (I + \alpha A)^\top (I + \alpha A) x \nonumber \\
&= x^\top \!\left(I + \alpha(A^\top + A) + \alpha^2 A^\top A\right) x \nonumber \\
&= \|x\|_2^2 + \alpha^2 \|Ax\|_2^2 ,
\label{eq:norm_expansion}
\end{align}
where the first-order term vanishes identically due to $A^\top + A = 0$. Using $\|Ax\|_2 \le \|A\|_2 \|x\|_2$, we obtain
\begin{equation}
\|x^{+}\|_2 \le \sqrt{1 + \alpha^2 \|A\|_2^2}\, \|x\|_2
\approx \left(1 + \tfrac{1}{2}\alpha^2 \|A\|_2^2\right)\|x\|_2 .
\label{eq:norm_bound}
\end{equation}
Thus, the norm variation induced by a single MCL update is of second order in $\alpha$. The same conclusion extends to matrix-valued hidden states $z \in \mathbb{R}^{C \times N}$ under the Frobenius norm.

\pb{Multi-Layer Composition and Growth Bound.}
For a sequence of updates
\begin{equation}
z_{l+1} = (I + \alpha A_l) z_l,
\qquad A_l^\top = -A_l, \quad l=0,\dots,L-1,
\label{eq:multi_step}
\end{equation}
the bound in Eq.~\eqref{eq:norm_bound} yields
\begin{equation}
\|z_L\|_2
\le
\left(\prod_{l=0}^{L-1} \sqrt{1 + \alpha^2 \|A_l\|_2^2}\right)\|z_0\|_2 .
\label{eq:product_bound}
\end{equation}
If $\|A_l\|_2 \le M$ for all $l$, then
\begin{equation}
\|z_L\|_2
\le
(1 + \alpha^2 M^2)^{L/2}\|z_0\|_2
\approx
\exp\!\left(\tfrac{L}{2}\alpha^2 M^2\right)\|z_0\|_2 .
\label{eq:exponential_bound}
\end{equation}
This shows that hidden-state growth is governed by $O(\alpha^2 L)$ rather than $O(\alpha L)$, providing a theoretical explanation for the enhanced stability of deep and long-horizon compositions.

\pb{Invertibility of Near-Isometric Updates.}
Finally, the linearized update $T = I + \alpha A$ possesses a well-defined inverse under standard step-size conditions.

\begin{proposition}[Invertibility of Near-Isometric Updates]
Let $A^\top = -A$ and $\alpha \|A\|_2 < 1$. Then $T = I + \alpha A$ is invertible, with inverse given by the Neumann series
\begin{equation}
T^{-1} = \sum_{k=0}^{\infty} (-\alpha A)^k ,
\label{eq:neumann_series}
\end{equation}
and satisfies the operator norm bound
\begin{equation}
\|T^{-1}\|_2 \le \frac{1}{1 - \alpha \|A\|_2}.
\label{eq:inverse_bound}
\end{equation}
\end{proposition}

\pb{Summary.}
MCL elevates hidden-state updates in neural operators from unconstrained Euclidean residuals to Lie-algebra-driven near-isometric evolution. The skew-symmetric structure eliminates first-order norm drift, leaving only second-order variations and yielding improved numerical stability under deep stacking and long-horizon rollout, while remaining computationally efficient and compatible with existing operator architectures.

\subsection{Baseline Models}\label{app:model}
\pb{Fourier Neural Operator.}
Fourier Neural Operator (FNO)~\cite{li2020fourier} is the seminal work in neural operator learning. 
Instead of directly learning the mapping on discrete grids, it learns the mapping between function spaces. 
To learn the mapping of partial differential equations, FNO utilizes the Fourier transform to convert complex global convolution operations into simple pointwise multiplications in the frequency domain. 
FNO consists of a boosting layer, followed by four cascaded spectral convolution layers, and finally outputs the results through two projection layers. 
In FNO, we retain 16 low-frequency modes and set the channel dimension to 64.
We implement FNO by transforming the input features via a spectral convolution layer.
Specifically, we convert to frequency domain with the Fourier transform, multiply it with learnable weight, and revert it back with the inverse Fourier transform.
We also add a residual connection via $1\times1$ convolution to preserve the high-frequency information.
We insert the MCL post activation function in every spectral convolution layer.
This position enables enforcing manifold constraints right after operation in the frequency domain, which enhances the stability of the model in long-term operation.

\pb{U-shaped Neural Operators.}
U-shaped Neural Operators (U-NO)~\cite{rahman2022u} addresses the issue that FNO tends to lose high-frequency details in deep networks. 
U-NO leverage U-Net architecture to perform downsampling and upsampling operations between spectral convolution layers.
This enables U-NO to capture both macroscopic flows and fine structures (such as shock waves and turbulence) at different resolution scales and fuse them through skip connections.
We implement U-NO with a 3-layer U-Net structure with low-frequency modes set to 12 and an initial channel width set to 178 (further doubled with the increasing network depth).
We incorporate MCL after every spectral convolution and before the activation function to strengthen manifold consistency of multi-scale features as well as to improve the model's ability to preserve high-frequency details.

\pb{Geometry-Informed Neural Operator.}
Geometry-Informed Neural Operator (GINO)~\cite{li2020geometry} leverages a graph neural network (GNN) to improve FNO's flexibility in handling irregular geometries.
The \say{Graph-Grid-Graph} architecture in GINO maps unstructured data to a latent regular grid with Fast Fourier Transform to capture global dependencies.
GINO operates at $O(N \log N)$ complexity, solving the computational bottleneck of GNN in large-scale complex geometry simulations. 
In \say{graph-to-grid} operation, GINO performs geometric projection of locally enriched features from the unstructured inputs into a regular grid.
To avoid geometry loss, GINO leverages geometric feature embedding into the physical features.
The spectral block utilizes a pointwise linear transformation, followed by spectral convolution and the activation function.
Finally, \say{grid-to-graph} operation maps the grid back to the original point with a GNN decoder.
In GINO, we insert MCL after every spectral convolution and before the activation function.

\subsection{Main and Ablation Results}\label{app:results}
\begin{figure}[!htb]
    \centering
    \includegraphics[width=.9\linewidth]{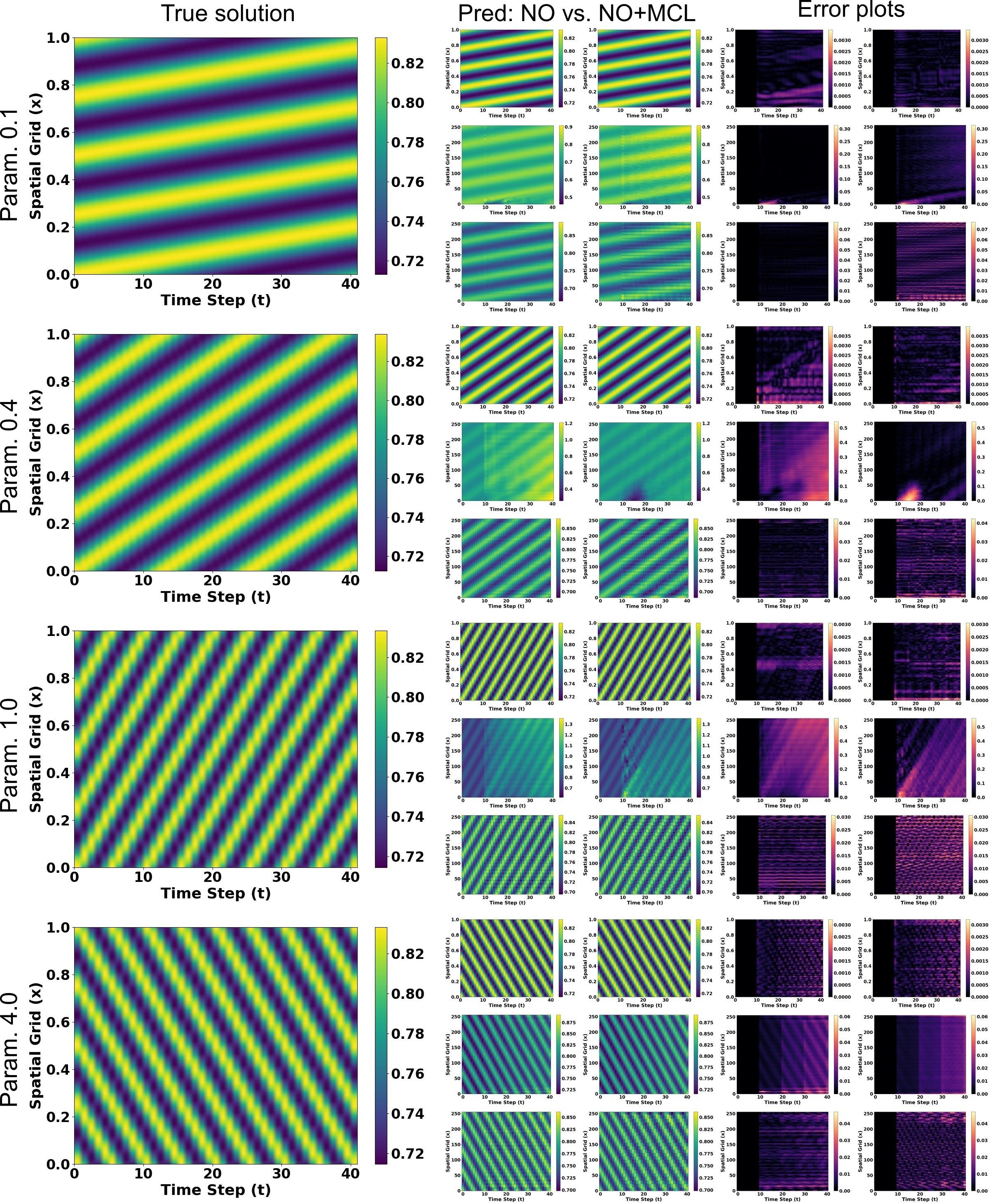}
    \caption{Plots on true solution, prediction on NO vs. NO+MCL, and the corresponding error plots on 1-D linear advection equation. Models used for prediction are, from top to bottom, FNO, U-NO, and GINO.}
    \label{fig:advection}
\end{figure}

\begin{figure}[!htb]
    \centering
    \includegraphics[width=.9\linewidth]{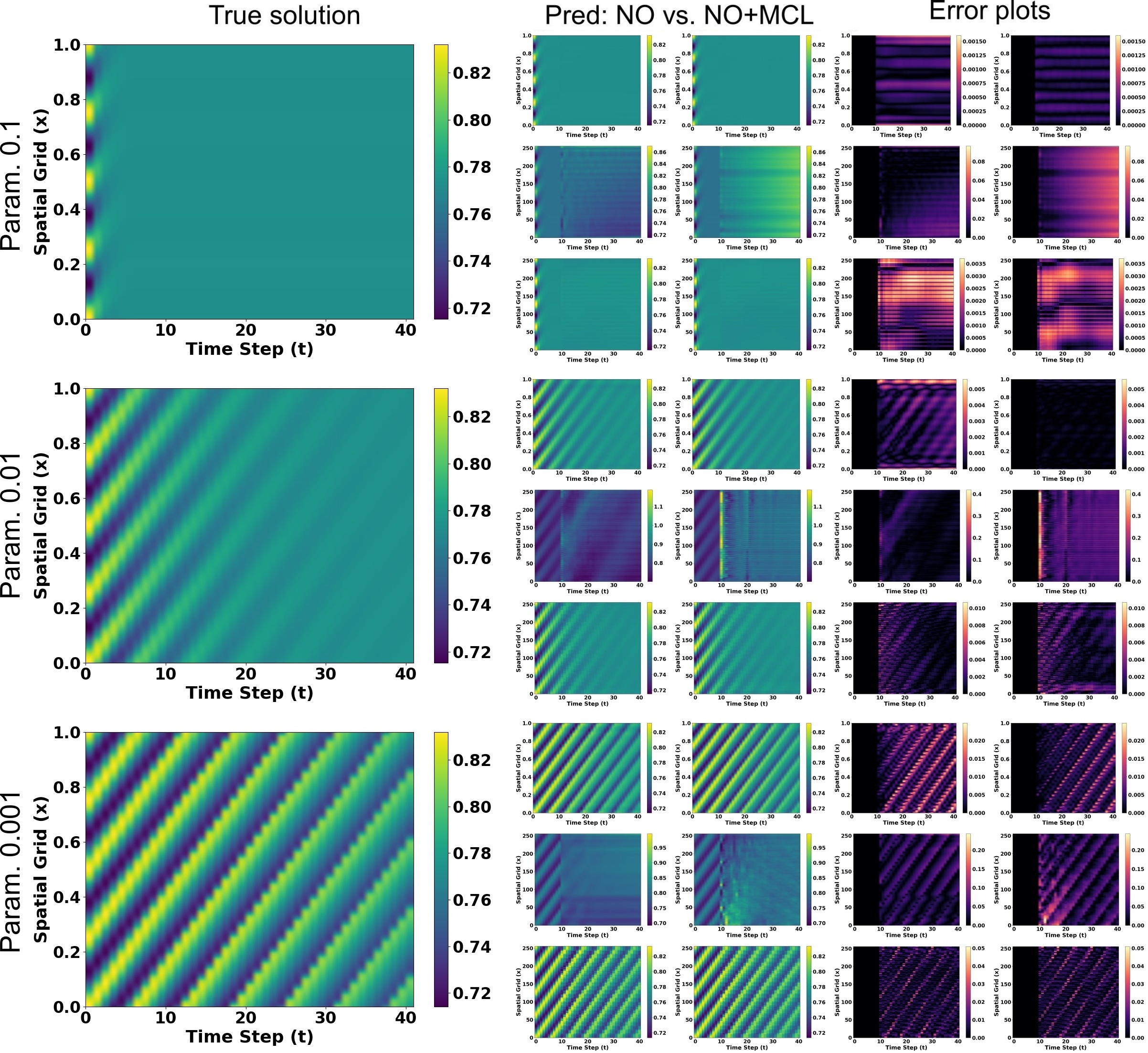}
    \caption{Plots on true solution, prediction on NO vs. NO+MCL, and the corresponding error plots on 1-D Burgers equation. Models used for prediction are, from top to bottom, FNO, U-NO, and GINO.}
    \label{fig:burgers}
\end{figure}

\begin{figure}[!htb]
    \centering
    \includegraphics[width=.9\linewidth]{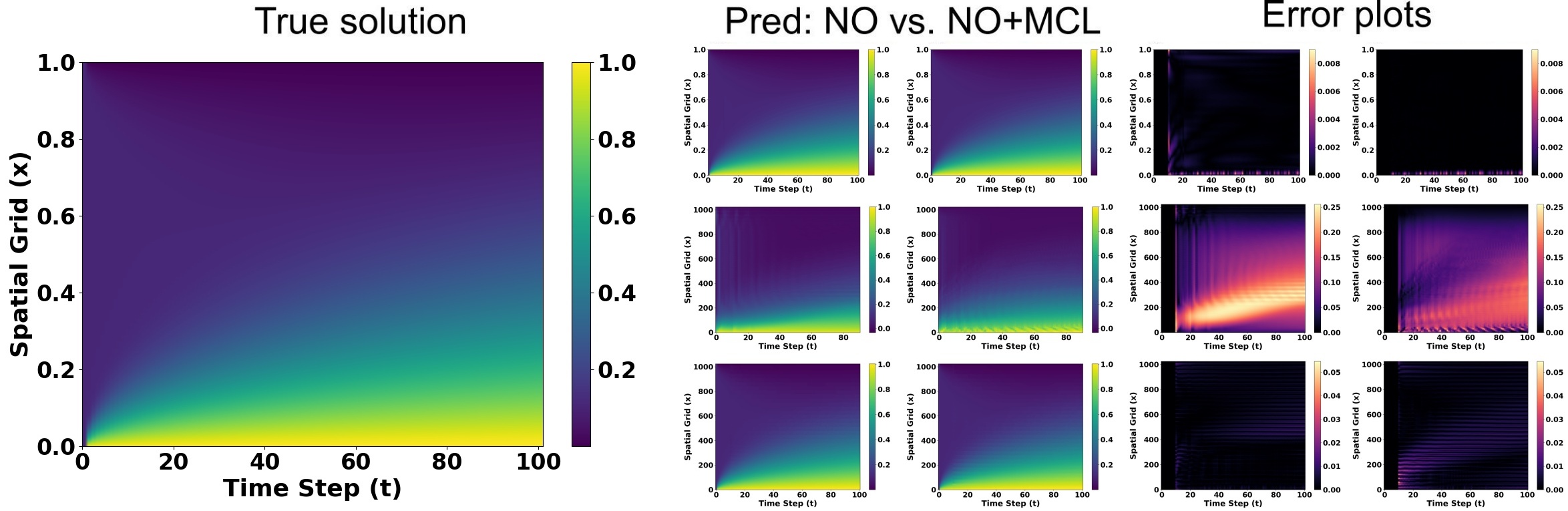}
    \caption{Plots on true solution, prediction on NO vs. NO+MCL, and the corresponding error plots on 1-D diffusion-sorption equation. Models used for prediction are, from top to bottom, FNO, U-NO, and GINO.}
    \label{fig:diffusion}
\end{figure}

\begin{figure}[!htb]
    \centering
    \includegraphics[width=.9\linewidth]{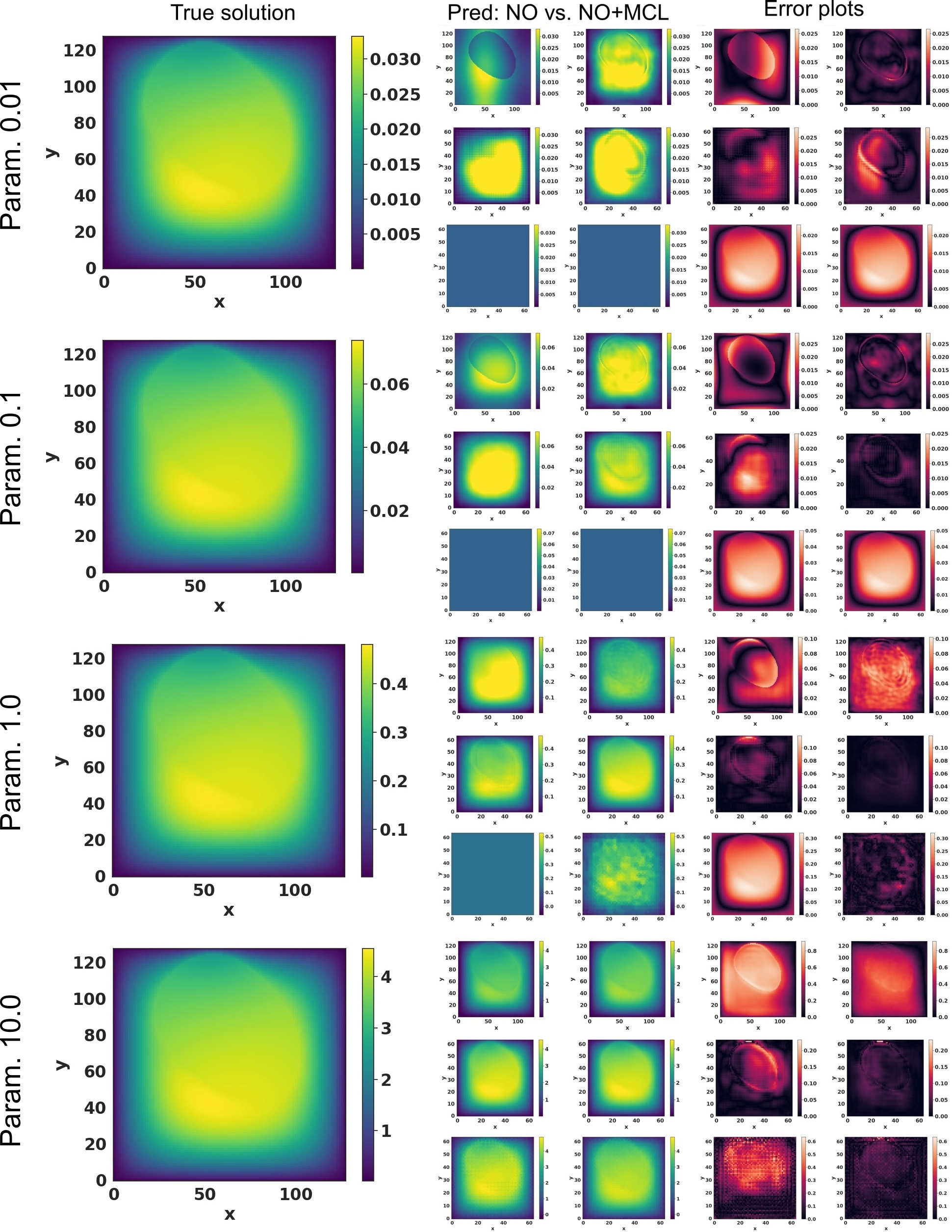}
    \caption{Plots on true solution, prediction on NO vs. NO+MCL, and the corresponding error plots on 2-D Darcy flow equation. Models used for prediction are, from top to bottom, FNO, U-NO, and GINO.}
    \label{fig:darcy}
\end{figure}

\begin{figure}[!htb]
    \centering
    \includegraphics[width=.9\linewidth]{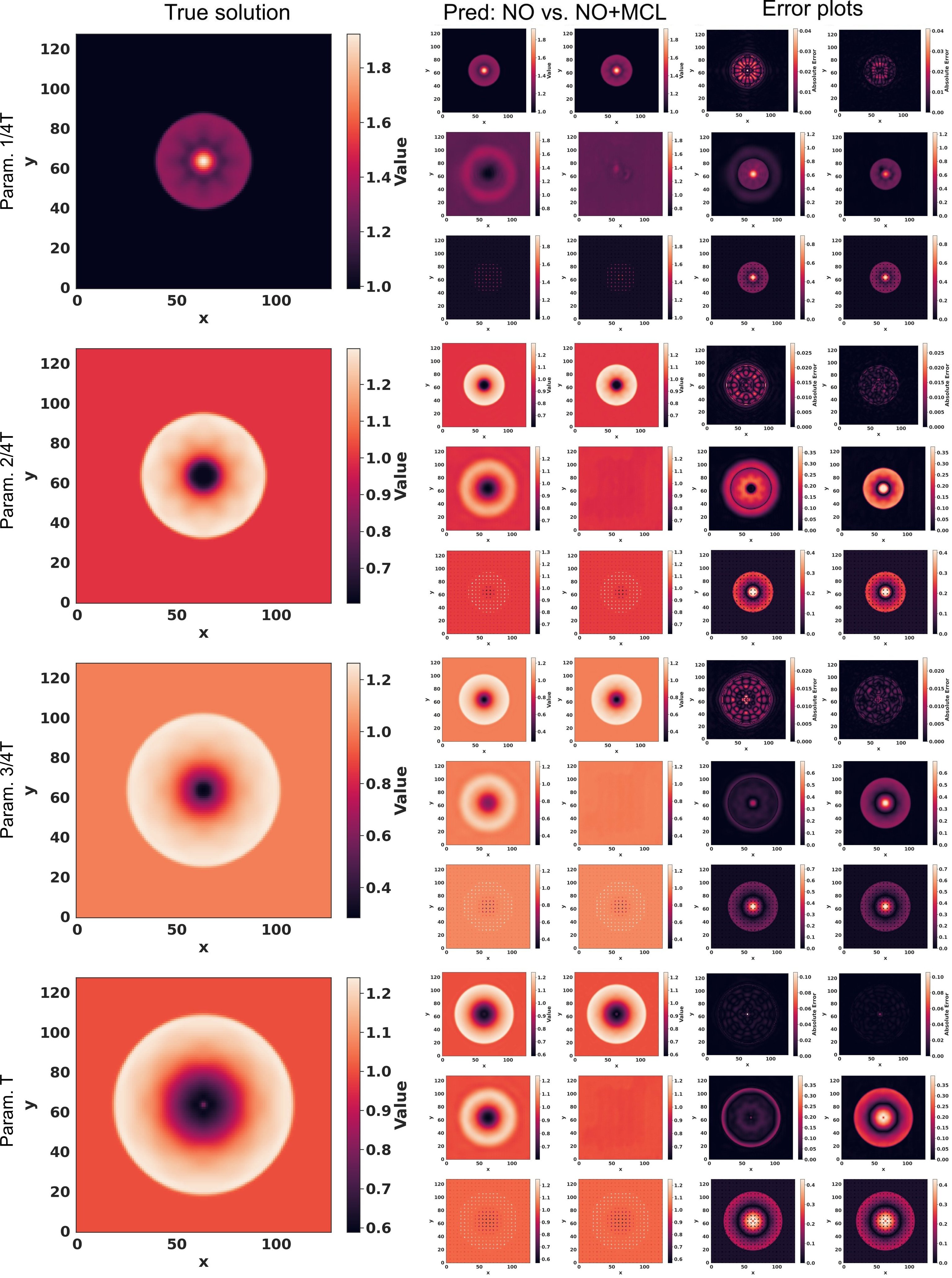}
    \caption{Plots on true solution, prediction on NO vs. NO+MCL, and the corresponding error plots on 2-D shallow-water equation. Models used for prediction are, from top to bottom, FNO, U-NO, and GINO.}
    \label{fig:shallow}
\end{figure}

\begin{figure}[!htb]
    \centering
    \includegraphics[width=.9\linewidth]{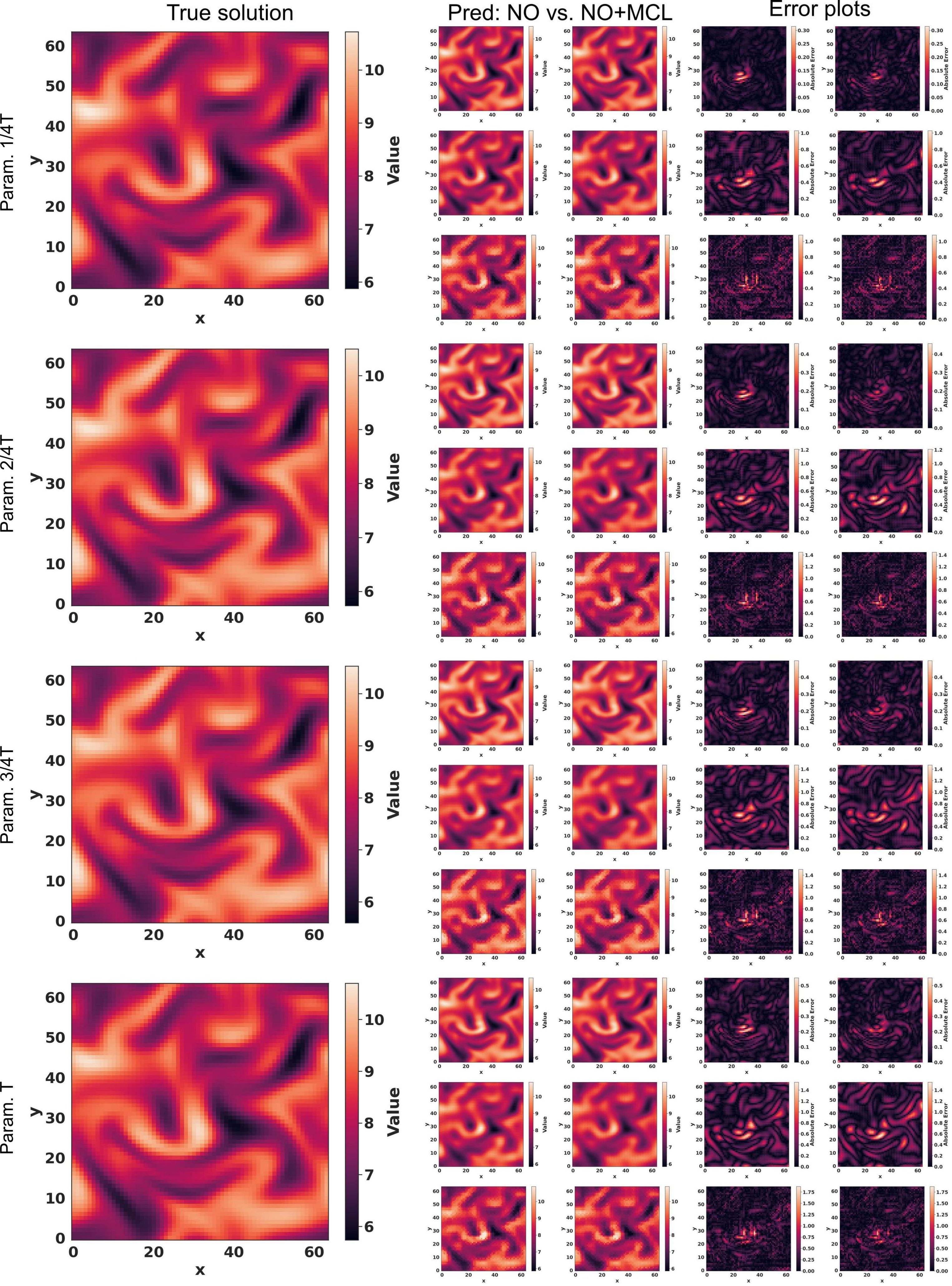}
    \caption{Plots on true solution, prediction on NO vs. NO+MCL, and the corresponding error plots on 2-D Navier-Stokes equation. Models used for prediction are, from top to bottom, FNO, U-NO, and GINO.}
    \label{fig:navier}
\end{figure}

We provide more results for the main experiment on 1-D Burgers equation (Table~\ref{tab:moreburgers}), 1-D linear advection equation (Table~\ref{tab:moreadvection}), 1-D diffusion-sorption equation (Table~\ref{tab:morediffusion}), 2-D Darcy flow equation (Table~\ref{tab:moredarcy}), 2-D shallow-water equation (Table~\ref{tab:moreshallow}), and 2-D Navier-Stokes (Table~\ref{tab:movenavier}) equation on various parameters and steps configurations.
Further, Table~\ref{tab:moreablate} provides more experiment on the contribution of MCL to improving the long-term prediction stability on neural operator across various parameters and step choices.
We also provide visualization on the comparison among true solution, neural operators (i.e., FNO, U-NO, and GINO) predictions, predictions with MCL integration, and the corresponding error plots for 1-D Burgers equation (Figure~\ref{fig:burgers}), 1-D linear advection equation (Figure~\ref{fig:advection}), 1-D diffusion-sorption equation (Figure~\ref{fig:diffusion}), 2-D Darcy flow equation (Figure~\ref{fig:darcy}), 2-D shallow-water equation (Figure~\ref{fig:shallow}), and 2-D Navier-Stokes (Figure~\ref{fig:navier}) equation.

\begin{table*}[!htb]
    \centering
    \caption{2-D Navier-Stokes}
    \label{tab:movenavier}
    \begin{tabular}{cccccccc}
        \toprule
        \multirow{2}*{Step} & \multirow{2}*{Model} &\multicolumn{3}{c}{Model only} & \multicolumn{3}{c}{Model + MCL}\\
        \cmidrule(lr){3-5} \cmidrule(lr){6-8}
        && MSE & Rel\_L2 & Rel\_H1 & MSE & Rel\_L2 & Rel\_H1\\
        \midrule
         \multirow{3}*{1} & FNO & 0.094958 & 0.006878 & 0.007343 & 0.003067 & 0.005678 & 0.006499\\
         & U-NO & 0.121584 & 0.090440 & 0.099027 & 0.117830 & 0.085402 & 0.094537\\
         & GINO & 0.014434 & 0.013638 & 0.021241 & 0.013268 & 0.013185 & 0.020583\\
         \midrule
         \multirow{3}*{3} & FNO & 0.002689 & 0.006452 & 0.007013 & 0.002607 & 0.005090 & 0.005856\\
         & U-NO & 0.073323 & 0.040074 & 0.047294 & 0.044330 & 0.037223 & 0.046833\\
         & GINO & 0.011692 & 0.012094 & 0.019547 & 0.010847 & 0.011832 & 0.019212\\
         \midrule
         \multirow{3}*{5} & FNO & 0.002618 & 0.006557 & 0.007242 & 0.002553 & 0.005167 & 0.006075\\
         & U-NO & 0.062031 & 0.044698 & 0.052281 & 0.044285 & 0.039946 & 0.046540\\
         & GINO & 0.011598 & 0.011943 & 0.019204 & 0.010699 & 0.011584 & 0.018829\\
         \midrule
         \multirow{3}*{10} & FNO & 0.003486 & 0.007163 & 0.008197 & 0.003328 & 0.005932 & 0.007190\\
         & U-NO & 0.064937 & 0.059068 & 0.064250 & 0.051550 & 0.051766 & 0.057728\\
         & GINO & 0.014704 & 0.013079 & 0.019682 & 0.012966 & 0.012962 & 0.019650\\
         \bottomrule      
    \end{tabular}
\end{table*}

\begin{table*}[!htb]
    \centering
    \caption{2-D Darcy Flow}
    \label{tab:moredarcy}
    \begin{tabular}{cccccccc}
        \toprule
        \multirow{2}*{Parameter} & \multirow{2}*{Model} &\multicolumn{3}{c}{Model only} & \multicolumn{3}{c}{Model + MCL}\\
        \cmidrule(lr){3-5} \cmidrule(lr){6-8}
        && MSE & Rel\_L2 & Rel\_H1 & MSE & Rel\_L2 & Rel\_H1\\
        \midrule
         \multirow{3}*{0.01} & FNO & 0.00057275 & 2.59687380 & 2.61742788 & 0.00019451 & 1.04014158 & 1.06347932\\
         & U-NO & 0.00022823 & 1.69248305 & 1.78475725 & 0.00021132 & 1.36236886 & 1.64780834\\
         & GINO & 0.00106853 & 5.07440402 & 5.05898289 & 0.00106853 & 5.06830841 & 5.05290873\\
         \midrule
         \multirow{3}*{0.1} & FNO & 0.00044092 & 0.44783878 & 0.45378846 & 0.00021056 & 0.24423277 & 0.25018481\\
         & U-NO & 0.00022993 & 0.33987113 & 0.34327716 & 0.00023840 & 0.15001997 & 0.16205981\\
         & GINO & 0.00161160 & 1.02423646 & 1.02470871 & 0.00161160 & 1.02382971 & 1.02430447\\
         \midrule
         \multirow{3}*{1.0} & FNO & 0.00085794 & 0.12478112 & 0.12827607 & 0.00030711 & 0.05464338 & 0.06032316\\
         & U-NO & 0.00036061 & 0.06873222 & 0.07676553 & 0.00026849 & 0.02492838 & 0.02671770\\
         & GINO & 0.01872887 & 0.67226677 & 0.67533560 & 0.00242101 & 0.24465922 & 0.28335634\\
         \midrule
         \multirow{3}*{10.0} & FNO & 0.00254715 & 0.02372490 & 0.02609825 & 0.00147270 & 0.01427597 & 0.01556954\\
         & U-NO & 0.00116923 & 0.01587893 & 0.01779863 & 0.00069153 & 0.00964605 & 0.01112830\\
         & GINO & 0.00762643 & 0.03921129 & 0.05773346 & 0.00349587 & 0.03141511 & 0.05131955\\
         \bottomrule      
    \end{tabular}
\end{table*}

\begin{table*}[!htb]
    \centering
    \caption{1-D Burgers}
    \label{tab:moreburgers}
    \begin{tabular}{ccccccccc}
        \toprule
        \multirow{2}*{Parameter} & \multirow{2}*{Step} & \multirow{2}*{Model} &\multicolumn{3}{c}{Model only} & \multicolumn{3}{c}{Model + MCL}\\
        \cmidrule(lr){4-6} \cmidrule(lr){7-9}
        &&& MSE & Rel\_L2 & Rel\_H1 & MSE & Rel\_L2 & Rel\_H1\\
        \midrule
         \multirow{12}*{0.1} & \multirow{3}*{1} & FNO & 0.000010 & 0.006182 & 0.006210 & 0.000001 & 0.002284 & 0.002361\\
         && U-NO & 0.033765 & 0.452617 & 0.476068 & 0.002880 & 0.099611 & 0.104958\\
         && GINO & 0.000219 & 0.039598 & 0.043709 & 0.000198 & 0.038372 & 0.042604\\
         \cmidrule(lr){2-9}
         & \multirow{3}*{10} & FNO & 0.000002 & 0.003629 & 0.003636 & 0.000001 & 0.001334 & 0.001347 \\
         && U-NO & 0.005832 & 0.205232 & 0.225318 & 0.001769 & 0.067778 & 0.070619\\
         && GINO & 0.000079 & 0.025896 & 0.027623 & 0.000067 & 0.024902 & 0.026560\\
         \cmidrule(lr){2-9}
         & \multirow{3}*{25} & FNO & 0.000001 & 0.004165 & 0.004170 & 0.000001 & 0.001703 & 0.001708 \\
         && U-NO & 0.002194 & 0.200498 & 0.216083 & 0.002786 & 0.092864 & 0.094666\\
         && GINO & 0.000035 & 0.024586 & 0.025107 & 0.000028 & 0.024180 & 0.024751\\
         \cmidrule(lr){2-9}
         & \multirow{3}*{30} & FNO & 0.000001 & 0.004510 & 0.004514 & 0.000001 & 0.001845 & 0.001851\\
         && U-NO & 0.002010 & 0.188081 & 0.204366 & 0.003395 & 0.102793 & 0.104488\\
         && GINO & 0.000031 & 0.025909 & 0.026262 & 0.000025 & 0.025021 & 0.025422\\
         \midrule

        \multirow{12}*{0.01} & \multirow{3}*{1} & FNO & 0.000113 & 0.014811 & 0.016098 & 0.000020 & 0.008317 & 0.008839\\
         && U-NO & 0.165838 & 1.069344 & 1.085058 & 0.043527 & 0.474051 & 0.524938\\
         && GINO & 0.001209 & 0.076995 & 0.100792 & 0.001180 & 0.076260 & 0.099925\\
         \cmidrule(lr){2-9}
         & \multirow{3}*{10} & FNO & 0.000023 & 0.011790 & 0.012122 & 0.000003 & 0.004339 & 0.004482\\
         && U-NO & 0.026230 & 0.446244 & 0.454794 & 0.017985 & 0.392924 & 0.412646\\
         && GINO & 0.000324 & 0.048874 & 0.064819 & 0.000324 & 0.048874 & 0.064819\\
         \cmidrule(lr){2-9}
         & \multirow{3}*{25} & FNO & 0.000007 & 0.008738 & 0.008843 & 0.000001 & 0.003393 & 0.003449\\
         && U-NO & 0.010297 & 0.423201 & 0.438888 & 0.016554 & 0.388093 & 0.393161\\
         && GINO & 0.000100 & 0.034658 & 0.044444 & 0.000099 & 0.033750 & 0.043469\\
         \cmidrule(lr){2-9}
         & \multirow{3}*{30} & FNO & 0.000006 & 0.008894 & 0.008979 & 0.000001 & 0.003339 & 0.003389\\
         && U-NO & 0.010876 & 0.448066 & 0.463097 & 0.016985 & 0.397757 & 0.402178\\
         && GINO & 0.000075 & 0.032197 & 0.040660 & 0.000074 & 0.031247 & 0.039604\\
         \midrule

        \multirow{12}*{0.001} & \multirow{3}*{1} & FNO & 0.000550 & 0.046626 & 0.068925 & 0.000327 & 0.035689 & 0.049867\\
         && U-NO & 0.042886 & 0.398778 & 0.415302 & 0.039507 & 0.350689 & 0.367253\\
         && GINO & 0.003950 & 0.147548 & 0.192726 & 0.003891 & 0.146621 & 0.192288\\
         \cmidrule(lr){2-9}
         & \multirow{3}*{10} & FNO & 0.000214 & 0.037893 & 0.050876 & 0.000108 & 0.026429 & 0.034554\\
         && U-NO & 0.025308 & 0.412722 & 0.417508 & 0.020423 & 0.398555 & 0.404843\\
         && GINO & 0.001475 & 0.109261 & 0.139855 & 0.001446 & 0.107881 & 0.138665\\
         \cmidrule(lr){2-9}
         & \multirow{3}*{25} & FNO & 0.000098 & 0.032601 & 0.039601 & 0.000046 & 0.022443 & 0.026869\\
         && U-NO & 0.014334 & 0.373097 & 0.376473 & 0.008402 & 0.331120 & 0.334335\\
         && GINO & 0.000586 & 0.082725 & 0.102302 & 0.000573 & 0.081434 & 0.101060\\
         \cmidrule(lr){2-9}
         & \multirow{3}*{30} & FNO & 0.000087 & 0.032831 & 0.038601 & 0.000040 & 0.022475 & 0.026118\\
         && U-NO & 0.013494 & 0.374821 & 0.377875 & 0.007504 & 0.325643 & 0.328482\\
         && GINO & 0.000485 & 0.079814 & 0.097431 & 0.000468 & 0.078492 & 0.095980\\
         \bottomrule      
    \end{tabular}
\end{table*}

\begin{table*}[!htb]
    \centering
    \small
    \caption{1-D Advection}
    \label{tab:moreadvection}
    \begin{tabular}{ccccccccc}
        \toprule
        \multirow{2}*{Parameter} & \multirow{2}*{Step} & \multirow{2}*{Model} &\multicolumn{3}{c}{Model only} & \multicolumn{3}{c}{Model + MCL}\\
        \cmidrule(lr){4-6} \cmidrule(lr){7-9}
        &&& MSE & Rel\_L2 & Rel\_H1 & MSE & Rel\_L2 & Rel\_H1\\
        \midrule
         \multirow{12}*{0.1} & \multirow{3}*{1} & FNO & 0.000037 & 0.005690 & 0.006920 & 0.000016 & 0.004433 & 0.005201\\
         && U-NO & 0.002880 & 0.099611 & 0.104958 & 0.002003 & 0.069950 & 0.079686\\
         && GINO & 0.002394 & 0.060020 & 0.069219 & 0.001251 & 0.038808 & 0.042497\\
         \cmidrule(lr){2-9}
         & \multirow{3}*{10} & FNO & 0.000096 & 0.009490 & 0.010656 & 0.000021 & 0.004975 & 0.005685\\
         && U-NO & 0.005685 & 0.07778 & 0.070619 & 0.001110 & 0.052989 & 0.059960\\
         && GINO & 0.002192 & 0.055833 & 0.065054 & 0.000817 & 0.031625 & 0.035272\\
         \cmidrule(lr){2-9}
         & \multirow{3}*{25} & FNO & 0.000131 & 0.011302 & 0.012323 & 0.000031 & 0.006246 & 0.006954\\
         && U-NO & 0.002786 & 0.092864 & 0.094666 & 0.001944 & 0.070169 & 0.075670\\
         && GINO & 0.003237 & 0.069133 & 0.077504 & 0.000879 & 0.033464 & 0.036768\\
         \cmidrule(lr){2-9}
         & \multirow{3}*{30} & FNO & 0.000148 & 0.012186 & 0.013224 & 0.000037 & 0.006938 & 0.007660\\
         && U-NO & 0.003395 & 0.102793 & 0.104488 & 0.002532 & 0.079815 & 0.084829\\
         && GINO & 0.003787 & 0.073401 & 0.081492 & 0.000947 & 0.034805 & 0.038071\\
         \midrule

         \multirow{12}*{0.4} & \multirow{3}*{1} & FNO & 0.000105 & 0.010102 & 0.011377 & 0.000044 & 0.007285 & 0.008363\\
         && U-NO & 0.009737 & 0.160042 & 0.166682 & 0.003026 & 0.075842 & 0.081295\\
         && GINO & 0.001250 & 0.040377 & 0.045325 & 0.001038 & 0.036694 & 0.040558\\
         \cmidrule(lr){2-9}
         & \multirow{3}*{10} & FNO & 0.000229 & 0.014032 & 0.015404 & 0.000059 & 0.008387 & 0.009435\\
         && U-NO & 0.031579 & 0.241814 & 0.244939 & 0.018542 & 0.179830 & 0.182581\\
         && GINO & 0.001341 & 0.040800 & 0.045707 & 0.001038 & 0.038424 & 0.042136\\
         \cmidrule(lr){2-9}
         & \multirow{3}*{25} & FNO & 0.000185 & 0.013393 & 0.014674 & 0.000059 & 0.008892 & 0.009906\\
         && U-NO & 0.036163 & 0.265120 & 0.267467 & 0.005913 & 0.122513 & 0.125383\\
         && GINO & 0.001659 & 0.045642 & 0.050466 & 0.001486 & 0.042745 & 0.046224\\
         \cmidrule(lr){2-9}
         & \multirow{3}*{30} & FNO & 0.000208 & 0.014030 & 0.015409 & 0.000070 & 0.009612 & 0.010718\\
         && U-NO & 0.044961 & 0.292901 & 0.295145 & 0.006926 & 0.132272 & 0.134956\\
         && GINO & 0.001808 & 0.047572 & 0.052251 & 0.001600 & 0.045318 & 0.048681\\
         \midrule

         \multirow{12}*{1.0} & \multirow{3}*{1} & FNO & 0.000243 & 0.013964 & 0.015500 & 0.000064 & 0.008839 & 0.009975\\
         && U-NO & 0.013008 & 0.189304 & 0.195135 & 0.012668 & 0.174245 & 0.178913\\
         && GINO & 0.003759 & 0.074602 & 0.089012 & 0.001234 & 0.038764 & 0.045873\\
         \cmidrule(lr){2-9}
         & \multirow{3}*{10} & FNO & 0.000258 & 0.014164 & 0.015734 & 0.000062 & 0.008699 & 0.009874\\
         && U-NO & 0.042200 & 0.309229 & 0.311331 & 0.018882 & 0.188371 & 0.190818 \\
         && GINO & 0.004167 & 0.077450 & 0.091834 & 0.001279 & 0.039344 & 0.046414\\
         \cmidrule(lr){2-9}
         & \multirow{3}*{25} & FNO & 0.000217 & 0.014444 & 0.016174 & 0.000060 & 0.009033 & 0.010409\\
         && U-NO & 0.042698 & 0.289166 & 0.290737 & 0.019341 & 0.187799 & 0.189771\\
         && GINO & 0.004569 & 0.081369 & 0.095214 & 0.001297 & 0.039893 & 0.046892\\
         \cmidrule(lr){2-9}
         & \multirow{3}*{30} & FNO & 0.000232 & 0.014673 & 0.016432 & 0.000067 & 0.009505 & 0.010917\\
         && U-NO & 0.053394 & 0.320988 & 0.322263 & 0.022905 & 0.204069 & 0.205847\\
         && GINO & 0.004382 & 0.080636 & 0.094342 & 0.001254 & 0.039865 & 0.046753\\
         \midrule

         \multirow{12}*{4.0} & \multirow{3}*{1} & FNO & 0.000041 & 0.008642 & 0.009577 & 0.000038 & 0.007885 & 0.009180\\
         && U-NO & 0.000215 & 0.024796 & 0.030143 & 0.000123 & 0.018545 & 0.022414\\
         && GINO & 0.003402 & 0.059406 & 0.066515 & 0.001183 & 0.040688 & 0.045241\\
         \cmidrule(lr){2-9}
         & \multirow{3}*{10} & FNO & 0.000020 & 0.005352 & 0.006604 & 0.000009 & 0.003792 & 0.004845\\
         && U-NO & 0.000210 & 0.000210 & 0.031569 & 0.000079 & 0.014824 & 0.018662\\
         && GINO & 0.001169 & 0.039349 & 0.044207 & 0.001169 & 0.039349 & 0.044207\\
         \cmidrule(lr){2-9}
         & \multirow{3}*{25} & FNO & 0.000070 & 0.010550 & 0.012474 & 0.000024 & 0.006658 & 0.008233\\
         && U-NO & 0.000908 & 0.050214 & 0.054475 & 0.000417 & 0.033138 & 0.035685\\
         && GINO & 0.001745 & 0.049242 & 0.056358 & 0.001302 & 0.042601 & 0.047080\\
         \cmidrule(lr){2-9}
         & \multirow{3}*{30} & FNO & 0.000087 & 0.011952 & 0.013937 & 0.000031 & 0.007564 & 0.009179\\
         && U-NO & 0.001034 & 0.053469 & 0.057611 & 0.000471 & 0.035188 & 0.037643\\
         && GINO & 0.001653 & 0.049062 & 0.056127 & 0.001336 & 0.044437 & 0.048821\\
         \bottomrule      
    \end{tabular}
\end{table*}

\begin{table*}[!htb]
    \centering
    \normalsize
    \caption{1-D Diffusion}
    \label{tab:morediffusion}
    \begin{tabular}{cccccccc}
        \toprule
        \multirow{2}*{Step} & \multirow{2}*{Model} &\multicolumn{3}{c}{Model only} & \multicolumn{3}{c}{Model + MCL}\\
        \cmidrule(lr){3-5} \cmidrule(lr){6-8}
        && MSE & Rel\_L2 & Rel\_H1 & MSE & Rel\_L2 & Rel\_H1\\
        \midrule
         \multirow{3}*{1} & FNO & 0.000013 & 0.013330 & 0.013673 & 0.000000 & 0.001321 & 0.002699\\
         & U-NO & 0.016053 & 0.444866 & 0.447889 & 0.005528 & 0.264650 & 0.269736\\
         & GINO & 0.269736 & 0.024395 & 0.025905 & 0.000138 & 0.043518 & 0.048344\\
         \midrule
         \multirow{3}*{10} & FNO & 0.000001 & 0.002817 & 0.003851 & 0.000000 & 0.001211 & 0.002637\\
         & U-NO & 0.007221 & 0.275725 & 0.276714 & 0.004055 & 0.190123 & 0.190769\\
         & GINO & 0.000007 & 0.008046 & 0.008046 & 0.000043 & 0.021212 & 0.024184\\
         \midrule
         \multirow{3}*{25} & FNO & 0.000001 & 0.001751 & 0.002732 & 0.000000 & 0.001016 & 0.002226\\
         & U-NO & 0.015116 & 0.346884 & 0.347008 & 0.007619 & 0.235791 & 0.236084\\
         & GINO & 0.000005 & 0.005968 & 0.007054 & 0.000028 & 0.014957 & 0.017041\\
         \midrule
         \multirow{3}*{30} & FNO & 0.000001 & 0.001567 & 0.002451 & 0.000000 & 0.000908 & 0.001985\\
         & U-NO & 0.015927 & 0.345835 & 0.345900 & 0.008158 & 0.237223 & 0.237605\\
         & GINO & 0.000006 & 0.006414 & 0.007381 & 0.000026 & 0.013854 & 0.015759\\
         \bottomrule      
    \end{tabular}
\end{table*}

\begin{table*}[!htb]
    \centering
    \caption{2-D Shallow Water}
    \label{tab:moreshallow}
    \begin{tabular}{cccccccc}
        \toprule
        \multirow{2}*{Step} & \multirow{2}*{Model} &\multicolumn{3}{c}{Model only} & \multicolumn{3}{c}{Model + MCL}\\
        \cmidrule(lr){3-5} \cmidrule(lr){6-8}
        && MSE & Rel\_L2 & Rel\_H1 & MSE & Rel\_L2 & Rel\_H1\\
        \midrule
        \multirow{3}*{1} & FNO & 0.000021 & 0.004370 & 0.008327 & 0.000010 & 0.003084 & 0.005210 \\
         & U-NO & 0.035991 & 0.172072 & 0.172072 & 0.012362 & 0.101253 & 0.112789\\
         & GINO & 0.020287 & 0.131313 & 0.149931 & 0.020275 & 0.131233 & 0.150866\\
         \midrule
         \multirow{3}*{10} & FNO & 0.000015 & 0.003681 & 0.005530 & 0.000004 & 0.002056 & 0.003120\\
         & U-NO & 0.021521 & 0.137780 & 0.141981 & 0.016664 & 0.117650 & 0.122598\\
         & GINO & 0.014786 & 0.110963 & 0.127245 & 0.014782 & 0.110899 & 0.127448\\
         \midrule
         \multirow{3}*{25} & FNO & 0.000010 & 0.003055 & 0.004681 & 0.000003 & 0.001657 & 0.002351\\
         & U-NO & 0.020450 & 0.136177 & 0.138663 & 0.009696 & 0.089772 & 0.092683\\
         & GINO & 0.008818 & 0.086397 & 0.098830 & 0.008813 & 0.086314 & 0.098691\\
         \midrule
         \multirow{3}*{30} & FNO & 0.000009 & 0.002962 & 0.004512 & 0.000002 & 0.001570 & 0.002197\\
         & U-NO & 0.018355 & 0.129456 & 0.131830 & 0.009735 & 0.091528 & 0.094194\\
         & GINO & 0.007941 & 0.083209 & 0.094984 & 0.007935 & 0.083135 & 0.094958\\
         \bottomrule      
    \end{tabular}
\end{table*}

\begin{table*}[!htb]
    \centering
    \caption{Ablation on 1-D Advection}
    \label{tab:moreablate}
   \begin{tabular}{ccccccccc}
        \toprule
        \multirow{2}*{Parameter} & \multirow{2}*{Step} &\multicolumn{3}{c}{FNO + MCL} & \multicolumn{3}{c}{FNO + MLP}\\
        \cmidrule(lr){3-5} \cmidrule(lr){6-8}
        && MSE & Rel\_L2 & Rel\_H1 & MSE & Rel\_L2 & Rel\_H1\\
        \midrule

        \multirow{4}{*}{0.1} & 1 & 0.000016 & 0.004433 & 0.005201 & 0.000019 & 0.004549 & 0.005406\\
         & 10 & 0.000021 & 0.004890 & 0.005646 & 0.000023 & 0.010975 & 0.010685\\
         & 25 & 0.000031 & 0.006246 & 0.006954 & 0.000034 & 0.008008 & 0.008762\\
         & 30 & 0.000037 & 0.006938 & 0.007660 & 0.000039 & 0.013526 & 0.014306\\
         \midrule
         \multirow{4}{*}{0.4} & 1 & 0.000044 & 0.007285 & 0.008363 & 0.000070 & 0.009049 & 0.010143\\
         & 10 & 0.000059 & 0.008387 & 0.009435 & 0.000099 & 0.010246 & 0.011365\\
         & 25 & 0.000059 & 0.008892 & 0.009906 & 0.000103 & 0.010614 & 0.011677\\
         & 30 & 0.000070 & 0.009612 & 0.010718 & 0.000123 & 0.011089 & 0.012255\\
         \midrule
         \multirow{4}{*}{1.0} & 1 & 0.000064 & 0.008839 & 0.009975 & 0.000078 & 0.009654 & 0.010835\\
         & 10 & 0.000062 & 0.008699 & 0.009874 & 0.000081 & 0.009650 & 0.010892\\
         & 25 & 0.000060 & 0.009033 & 0.010409 & 0.000074 & 0.009923 & 0.011380\\
         & 30 & 0.000067 & 0.009505 & 0.010917 & 0.000083 & 0.010557 & 0.012029\\
         \midrule
         \multirow{4}{*}{4.0} & 1 & 0.000038 & 0.007885 & 0.009180 & 0.000053 & 0.009762 & 0.010448\\
         & 10 & 0.000009 & 0.003792 & 0.004845 & 0.000010 & 0.004100 & 0.005137\\
         & 25 & 0.000024 & 0.006658 & 0.008233 & 0.000027 & 0.006888 & 0.008490\\
         & 30 & 0.000031 & 0.007564 & 0.009179 & 0.000034 & 0.007833 & 0.009483\\
        \bottomrule
         
    \end{tabular}
    
\end{table*}

\end{document}